%% file: main.tex
\documentclass[runningheads]{llncs}

\PassOptionsToPackage{dvipsnames,table}{xcolor}
\usepackage{eccv}

\usepackage{eccvabbrv}

\usepackage{graphicx}
\usepackage{booktabs}
\usepackage{siunitx}
\usepackage{array}
\usepackage{colortbl}
\usepackage{multirow}
\usepackage{makecell}
\usepackage{adjustbox}
\usepackage{wrapfig}
\usepackage{amssymb}
\usepackage{amsmath}
\usepackage{pifont}
\usepackage{bbding}
\usepackage{xfrac}
\usepackage{subcaption}
\usepackage{enumitem}
\usepackage{tcolorbox}
\usepackage{rotating}
\usepackage{float}
\usepackage[utf8]{inputenc}
\usepackage{longtable}
\definecolor{posblue}{RGB}{26, 115, 232} 
\definecolor{negred}{RGB}{217, 48, 37}  
\definecolor{recipeblue}{RGB}{90,140,250}
\definecolor{mossgreen}{RGB}{0,114,60} 
\definecolor{mimosa}{RGB}{247,181,87}
\definecolor{latexorange}{RGB}{255,128,0}
\newcolumntype{g}{>{\columncolor{gray!10}}r}
\newcolumntype{w}{>{\columncolor{white}}r}

\definecolor{col_color}{RGB}{242, 247, 250}

\makeatletter
\newcounter{blankfootnote}

\newcommand{\blfootnote}[1]{%
  \begingroup
  \stepcounter{blankfootnote}%
  \renewcommand{\thefootnote}{}%
  \@ifundefined{theHfootnote}{}{%
    \renewcommand{\theHfootnote}{blankfootnote.\arabic{blankfootnote}}%
  }%
  \footnotetext{#1}%
  \endgroup
}
\makeatother

\input{math_definition}
\usepackage[accsupp]{axessibility}  

\usepackage{hyperref}

\usepackage{orcidlink}

\usepackage{wrapfig}
\usepackage{mathtools}

\begin{document}

\title{Lessons and Open Questions from a Unified Study of Camera-Trap Species Recognition Over Time} 

\titlerunning{Camera-Trap Species Recognition Over Time}



\author{
Sooyoung Jeon$^*$\inst{1} \and
Hongjie Tian$^*$\inst{1} \and
Lemeng Wang$^*$\inst{1} \and
Zheda Mai$^\dagger$\inst{1} \and \\
Vidhi Bakshi\inst{1} \and
Jiacheng Hou\inst{1} \and
Ping Zhang\inst{1} \and
Arpita Chowdhury\inst{1} \and \\
Jianyang Gu\inst{1} \and
Wei-Lun Chao\inst{2}
}

\authorrunning{S. Jeon et al.}


\institute{$^1$The Ohio State University \quad $^2$Boston University\\ \url{https://jeonso0907.github.io/stream-trap/}}



\maketitle
\blfootnote{$^*$Equal contribution. $^\dagger$Corresponding author (mai.145@osu.edu).}

\vspace{-2.0em}
\input{sec/0_abstract}
\input{sec/1_intro_new}

\input{sec/2_related}
\input{sec/3_benchmark}
\input{sec/4_0_adapt_hard}

\input{sec/4_1_investigation}

\input{sec/5_accum}

\input{sec/6_end_user}
\input{sec/7_conclusion}

%
%
\bibliographystyle{splncs04}
\bibliography{main}

\input{sec/X_suppl_eccv}
\end{document}

%% file: math_definition.tex
\usepackage{amssymb}
\usepackage{amsmath,amsfonts}
\usepackage{amsopn}
\usepackage{bm} 
\usepackage{multirow}
\usepackage{flushend}
\usepackage{tabularx}

\usepackage{lipsum}

\newcommand{\vct}[1]{\boldsymbol{#1}} 

\newcommand{\ProbOpr}[1]{\mathbb{#1}}

\newcommand{\expect}[2]{%
\ifthenelse{\equal{#2}{}}{\ProbOpr{E}_{#1}}
{\ifthenelse{\equal{#1}{}}{\ProbOpr{E}\left[#2\right]}{\ProbOpr{E}_{#1}\left[#2\right]}}} 

\DeclareMathOperator{\argmax}{arg\,max}

\newcommand{\vtheta}{\vct{\theta}}

\newcommand{\vx}{{\vct{x}}}

\newcommand{\vw}{\vct{w}}

\newcommand{\eat}[1]{}



%% file: sec/0_abstract.tex
\begin{abstract}

Camera traps are crucial for large-scale biodiversity monitoring, yet accurate automated analysis remains challenging due to diverse deployment environments.  While the computer vision community has predominantly framed this challenge as cross-domain (\eg, cross-site) generalization, this perspective overlooks a primary challenge faced by ecological practitioners: maintaining reliable recognition at the \emph{fixed site over time}, where the dynamic nature of ecosystems introduces profound temporal shifts in both background and animal distributions. To bridge this gap, we present the first unified study of \textbf{camera-trap species recognition over time}. We introduce a realistic, large-scale benchmark comprising 546 camera traps with a streaming protocol that evaluates models over chronologically ordered intervals. Our end-user-centric study yields four key findings. (1) Biological foundation models (\eg, BioCLIP~2) underperform at numerous sites even in initial intervals, underscoring the necessity of site-specific adaptation. (2) Adaptation is challenging under realistic evaluation: when models are updated using past data and evaluated on \emph{future} intervals (mirrors real deployment lifecycles), naive adaptation can even degrade below zero-shot performance. (3) We identify two main drivers of this difficulty: severe class imbalance and pronounced temporal shift in both species distribution and backgrounds between consecutive intervals. (4) We find that effective integration of model-update and post-processing techniques can largely improve accuracy, though a gap from the upper bounds remains. Finally, we highlight critical open questions, such as predicting when zero-shot models will succeed at a new site and determining whether/when model updates are necessary. Together, our benchmark and analysis provide actionable deployment guidelines for ecological practitioners while establishing new directions for future research in vision and machine learning.

\keywords{Temporal Shift \and Adaptation \and Ecological Monitoring}
\end{abstract}

%% file: sec/1_intro_new.tex

\begin{figure}[t]
\small
\centering
\noindent\includegraphics[width=1\linewidth]{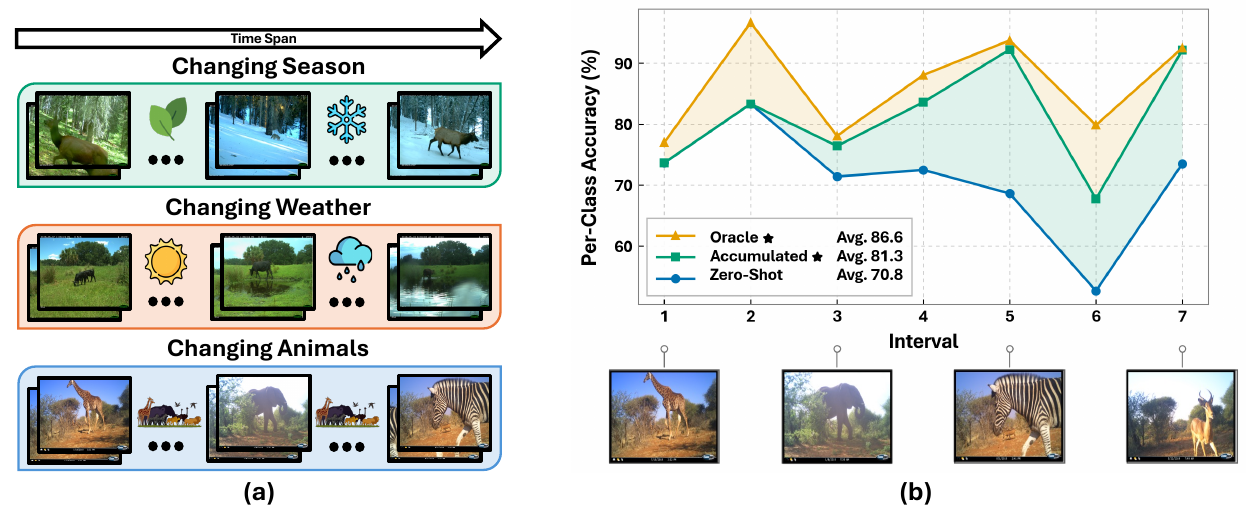}
\vspace{-5mm}
\caption{
\small
\textbf{(a). Temporal shifts:} Even at a fixed site, foreground and background appearances evolve over time due to changing seasons, weather, and animal distribution. \textbf{(b). Streaming evaluation: } Accumulated$\star$ model is fine-tuned with training data up to the current interval and evaluated on the test data of the next interval. For reference, an Oracle$\star$ model is fine-tuned with the union of training data from all time intervals at a site and evaluated on the test data of all intervals. ($\star$: our recipe \cref{sec:recipe}).
}
\label{fig:teaser}
\vspace{-0mm}
\end{figure}


\section{Introduction}
\label{sec:intro}

Camera traps are vital tools for ecology and conservation: they non-invasively capture large volumes of time-stamped images in natural habitats, supporting biodiversity monitoring, behavior analysis, and conservation planning~\cite{pollock2025harnessing,tuia2022perspectives}. Yet ecosystems are inherently non-stationary: seasonal shifts, weather changes, vegetation growth, and animal migration constantly reshape backgrounds and animal distributions (\cref{fig:teaser}a). Such dynamic conditions introduce substantial visual variability, posing unique challenges for vision models~\cite{koh2021wilds,beery2021iwildcam}.

Prior studies often frame the problem as domain adaptation or generalization, aiming to transfer models trained on source domains (\eg, specific cameras) to unseen target domains~\cite{sagawa2021extending,zhou2022domain}. While this perspective casts camera trap data as a crucial testbed for machine learning robustness, it diverges from the practical needs of end-users (\eg, ecologists and conservation practitioners). Their dominant interest is not a model's ranking on cross-domain leaderboards, but its \emph{ability to maintain high accuracy at a specific deployment site as dynamically changing data arrives}. Recognizing that this practical problem remains largely underexplored,  we conduct a unified and comprehensive study of \textbf{camera trap species recognition over time}, reformulating the evaluation to faithfully mirror the temporal lifecycle of real-world deployments.

To facilitate this study, we introduce a realistic camera trap benchmark \textsc{StreamTrap} with a streaming evaluation protocol where models are evaluated as new data arrives and are incrementally updated thereafter. For each camera trap, the image stream is partitioned into consecutive \textit{time intervals} for evaluating models chronologically. Concretely, at each interval $j$, a model can access cumulative, labeled training data from all intervals up to the $j$-th interval, undergo parameter updates, and evaluate on the unseen test data from the subsequent $(j+1)$-th interval. To construct \textsc{StreamTrap}, we curate data from the LILA BC repository~\cite{LILA-BC}, identifying \textbf{546} camera traps from 17 datasets, each with sufficient data spanning at least \textbf{6} months. Also, we provide a modular, reproducible data pipeline that adheres to FAIR principles (Findable, Accessible, Interoperable, and Reusable), enabling practitioners to seamlessly convert their raw ecological data into standardized temporal benchmarks.

Building on \textsc{StreamTrap}, we conduct a unified \emph{end-user-centric} study aiming to answer common questions faced by ecological practitioners, and distill actionable guidelines for maintaining accurate models in dynamic environments. We summarize our key analyses as follows:

\noindent\textbf{Adaptation is required even with biological foundation models.} Before deployment, end-users must decide whether a zero-shot biological vision foundation model is strong enough for a new camera trap, or whether site-specific adaptation is still needed. This decision is especially important as biological vision foundation models~\cite{gu2025bioclip,stevens2024bioclip,mai2026ava} already achieve state-of-the-art zero-shot accuracy across diverse biodiversity datasets, including subsets of the LILA BC repository. However, in \textsc{StreamTrap}, we observe substantial variability in BioCLIP 2's \textit{zero-shot} accuracy across 546 camera traps: while 161 traps exceed 90\% accuracy, demonstrating the promise of foundation models, another 162 fall in the 50--80\% range, indicating that site-specific adaptation remains critical for many deployment sites.

\noindent\textbf{Adaptation is challenging in the real world. } When off-the-shelf foundation models underperform at a deployment site, fine-tuning is a natural first adaptation choice~\cite{mai2024fine, mai2026revisiting, zhang2026adapting}. However, adaptation in \textsc{StreamTrap} is more challenging than standard in-domain fine-tuning: a model updated using data observed up to the current interval must be evaluated on the \emph{future} interval. Under this realistic streaming protocol, we surprisingly find that naively fine-tuning on all data observed so far (\textbf{accumulated model}) can degrade performance, often below the zero-shot baseline. Our investigation identifies two primary drivers behind this adaptation failure. First, as camera traps acquire images passively and must ``wait'' for animals to appear, some species may not be observed within a given interval. Thus, the training data available for an interval is often extremely low-shot and severely imbalanced, causing poor generalization during naive adaptation~\cite{tu2023holistic}. Second, due to the inherent non-stationarity of ecosystems, both the background context and the animal population distributions undergo profound shifts between consecutive intervals.  As a result, a model that achieves high accuracy on current and past intervals is not guaranteed to remain strong in the future intervals with shifted distributions.

\noindent\textbf{Practical recipes for robust adaptation. } To combat the low-shot and imbalance issues, we evaluate a suite of practical techniques and find that combining parameter-efficient fine-tuning~\cite{mai2025lessons} and class-imbalance losses~\cite{ren2020balanced,ye2020identifying} provides a robust recipe for camera-trap model adaptation. However, even with this improved recipe, continually adapted models still exhibit a non-negligible performance gap (\cref{fig:teaser}b) compared to an upper bound (\textbf{oracle$\star$} model fine-tuned with \emph{all} training data across all time intervals at a site). To further address the inter-interval distribution shifts and close this gap, we identify three post-processing techniques: post-hoc logit calibration~\cite{mai2024fine}, weight interpolation~\cite{wortsman2022robust}, and interval model selection.

\noindent\textbf{More practical considerations. } Accurate real-world deployment involves more than choosing an adaptation algorithm. Through discussions with end-users, we identify practical questions frequently overlooked by the vision community: \textit{How can practitioners predict if a foundation model's zero-shot performance will be sufficient at a new, unseen site? If not, must the model be updated at every interval, or is there an effective mechanism to determine when adapting at a given interval is actually necessary?} As these questions are largely underexplored, we provide preliminary analyses and baseline solutions.

\noindent \textbf{Contributions.} We present the first unified study of camera trap species recognition over time that reflects the temporal lifecycle of ecological monitoring. We systematically analyze critical deployment questions, providing actionable guidance on what works while urging the computer vision community to tackle the remaining open challenges. Our work serves two primary audiences:

\begin{itemize}[label=$\bullet$, topsep=0pt, partopsep=0pt, itemsep=0pt, parsep=0pt]
\item \textbf{End-users}: (1) We provide actionable guidance for leveraging foundation models and adaptation techniques in reliable camera-trap analysis; (2) Our FAIR-compliant data preparation pipeline helps practitioners convert raw ecological monitoring data into standardized, reusable benchmarks.

\item \textbf{Algorithm developers}: (1) We transform the LILA BC repository, which has historically been difficult for developers to use at scale, into a standardized benchmark for evaluating robust real-world adaptation.(2) We identify critical but overlooked challenges about \textit{what}, \textit{how}, and \textit{when} to update models under field constraints (\eg, scarce or imbalanced data, non-stationary streams). By steering development toward these practitioner-driven questions, we aim to accelerate progress in this vital scientific application area.
\end{itemize}

%% file: sec/2_related.tex
\section{Related Work}
\label{sec:related}
\noindent\textbf{Camera trap data in computer vision.} Camera traps are essential tools for biodiversity monitoring, producing large-scale wildlife image collections that reveal species richness and behavior~\cite{trolliet2014use, boitani2016camera}. To automate analysis, deep learning has been widely adopted for species detection and classification~\cite{norouzzadeh2018automatically, yu2013automated}. A major challenge is generalization: models trained at one location often fail when deployed at another. The iWildCam challenges~\cite{beery2019iwildcam, beery2021iwildcam} address this by splitting data by camera location to assess out-of-distribution generalization~\cite{koh2021wilds, mai2024fine}. Recent work further explores multimodal foundation models for richer camera-trap understanding~\cite{gabeff2024wildclip, fabian2023multimodal}. However, most prior work frames camera-trap analysis as a cross-domain generalization problem, whereas real deployments often require reliable species recognition at a \textbf{fixed site over time}, where seasonal changes, habitat dynamics, and animal migration continuously shift both species distributions and visual backgrounds, degrading model performance over long-term monitoring. We address this gap with a \textbf{unified, end-user-centric study} of camera-trap recognition across the temporal deployment lifecycle, characterizing adaptation challenges and deriving practical guidelines for long-term use.


\noindent\textbf{Continual learning (CL).}  CL studies how a model learns from a non-stationary data stream while retaining acquired knowledge and improving over time without catastrophic forgetting~\cite{mai2022online, li2026fasterpathcontinuallearning, mai2026survey}. This paradigm is closely related to our streaming evaluation protocol. However, standard CL assumes limited or no access to past data due to privacy or other constraints~\cite{cflat, mai2021supervised}.  In contrast, ecological deployments treat labeled camera trap data as a valuable resource that is archived and fully accessible. To isolate temporal shift effects and avoid conflating them with artificial storage constraints, our baseline uses all historically available data at each interval, corresponding to the cumulative training upper bound in CL. While \textsc{StreamTrap} is explicitly tailored for studying wildlife monitoring under temporal shifts, its chronologically ordered data streams and severe, natural distribution shifts inherently provide a challenging real-world testbed for the broader continual learning community.



\noindent\textbf{Class-imbalanced learning.}  
Camera trap datasets often follow a long-tailed distribution, where models perform well on the majority species but poorly on rare ones~\cite{bevan2024deep, malik2021two}. As rare species are often of greatest ecological interest, addressing this imbalance is critical. Common techniques can be broadly categorized into two groups: data-level methods modifying the data distribution by over(under)sampling; algorithm-level methods adjusting the learning process with specialized losses or algorithmic modifications~\cite{zhang2023deep, rezvani2023broad}. We provide more detailed related work in \cref{supp:related_work}.

%% file: sec/3_benchmark.tex
\section{\textsc{StreamTrap}: Benchmarking Real-World Camera Trap Deployment with Streaming Evaluation Protocol}
\label{sec: eval_proto}

\noindent\textbf{Motivation.} Camera traps are central to ecology and conservation: they non-invasively capture massive streams of time-stamped images in natural habitats, enabling biodiversity monitoring, behavior analysis, and conservation planning~\cite{pollock2025harnessing,tuia2022perspectives}. Yet the environments are inherently non-stationary: seasons, weather, vegetation, and migration continually reshape both the backgrounds and species distributions (\cref{fig:teaser}a), creating persistent visual shifts for deployed vision models. Prior work typically frames this challenge as domain adaptation / generalization, focusing on transferring models from source to target domains~\cite{sagawa2021extending,zhou2022domain}.  However, this framing does not fully capture the needs of end-users. They do not prioritize static cross-domain leaderboards; instead, their dominant interest is whether a model can \emph{maintain reliable accuracy at a fixed deployment site as new data arrive over time}.


Despite this practical need, existing benchmarks don't capture how camera-trap data arrive in the wild or reflect the temporal lifecycle of real deployments. To close this gap and refocus the community on deployment-critical evaluation, we introduce \textsc{StreamTrap}, a benchmark for camera trap species recognition over time with a streaming evaluation. By shifting the research focus from static domain transfer to long-term reliability, \textsc{StreamTrap} encourages the community to develop adaptive systems that remain accurate throughout real ecological monitoring deployments.

\begin{figure}[t]
\small
\centering
\noindent\includegraphics[width=1\linewidth]{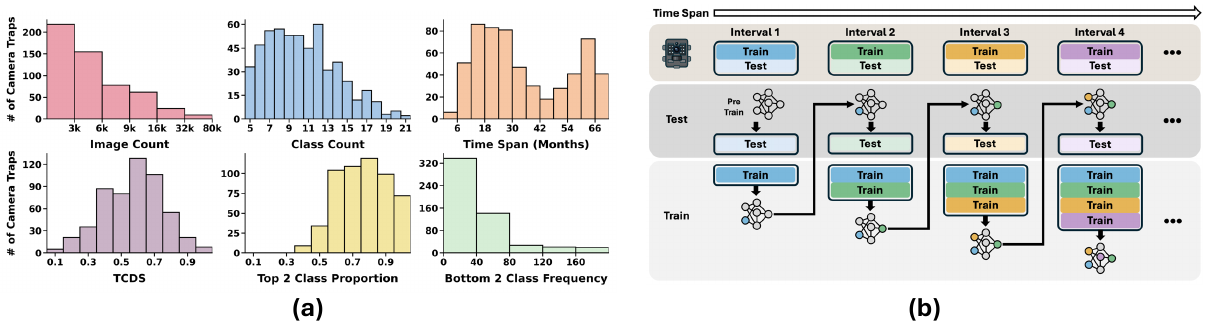}
\vspace{-5mm}
\caption{\small \textbf{(a). \textsc{StreamTrap} statistics}. Top row: histograms of the image, class, and time span; Bottom row: histograms of temporal shift (TCDS, defined in \cref{sec:investigation}) and class imbalance metrics (defined in \cref{ss:Fair Data Processing Pipeline}).   \textbf{(b). Streaming evaluation protocol: } models are continually fine-tuned on the labeled data observed so far and evaluated on future unseen data. }
\label{fig:stat_and_eval_protocal}
\vspace{-0mm}
\end{figure}

\noindent\textbf{Streaming evaluation protocol.} Unlike conventional domain adaptation, where target-domain data are often assumed to be available all at once~\cite{gong2012geodesic,singhal2023domain}, \textsc{StreamTrap} evaluates models under a realistic streaming protocol that better reflects camera-trap deployment: models are evaluated on future data as they arrive and may be updated after labeled data from previous intervals become available. Given time-stamped images from a camera trap, $\{(\vx_i, y_i, t_i)\}$, where $\vx_i$ denotes the $i$-th image, $y_i$ its label, and $t_i$ its timestamp, we partition the stream into a sequence of chronological intervals. At interval $j$, a model may access the accumulated labeled training data from intervals up to interval $j$, update its parameters, and is then evaluated on unseen test data from the subsequent $(j\!+\!1)$-th interval. \cref{fig:stat_and_eval_protocal}b illustrates this protocol.

Also, while \textsc{StreamTrap} is tailored for camera traps under temporal shift, its streaming formulation also serves as an excellent, highly challenging testbed for broader machine learning paradigms, such as online continual learning~\cite{mai2022online}.

\subsection{Benchmark Construction and Statistics}
\label{ss:Fair Data Processing Pipeline}

\noindent\textbf{Data source.} We build \textsc{StreamTrap} on the LILA BC repository~\cite{LILA-BC} with 17 camera-trap datasets (\eg, Snapshot Serengeti) comprising hundreds of camera traps deployed across continents. Here, a {camera trap} refers to a stationary camera installed at a fixed location to passively monitor wildlife activity.

\noindent\textbf{FAIR processing pipeline.} We focus on camera traps with sufficient temporal coverage and image volume to support streaming evaluation. We first partition each camera trap stream into 30-day intervals; intervals with fewer than 200 images are merged with adjacent intervals to ensure sufficient data for both training and testing. Within each interval, we split data into training and test sets, constructing the test split to be \textbf{class-balanced}. Species with fewer than 10 samples in an interval are held out as rare species to maintain evaluation stability: they are separately evaluated and analyzed in \cref{supp:relaxed_assumption}. To focus on the temporal shift problem, we follow the LILA BC protocol and use MegaDetector~\cite{beery2023megadetector} to extract image patches around detected animal instances. We enlarge each bounding box by 50\% to preserve background context. 

To promote the FAIR principles (Findable, Accessible, Interoperable, and Reusable) and enable future researchers to convert their raw ecological data into standardized benchmarks, we establish a modular data processing pipeline. Concretely, we preprocess each camera trap to ensure the availability of essential metadata (timestamps, image-level species labels, and animal bounding boxes) and apply two main filtering steps:

\begin{enumerate}[nosep,topsep=0pt,parsep=0pt,partopsep=0pt, leftmargin=*]
\item \textbf{Single-species filtering:} retain images containing a single species (no humans or vehicles) so that image-level labels can be unambiguously assigned to detected instances (removes $<2\%$ of images);
\item \textbf{Detection filtering:} retain images with at least one detected animal bounding box with confidence $>0.8$ to ensure reliable localization of animals.
\end{enumerate}
After filtering, we retain camera traps with more than {1,000} images spanning at least \textbf{6} months. Additional details, including design choices, preprocessing and pipeline steps, MegaDetector robustness checks, and the list of camera traps included, are provided in \cref{supp:benchmark}.

\noindent\textbf{Basic statistics.} 
\label{stat_metrics}
Our rigorous processing pipeline yielded 546 valid camera traps. A summary of statistics (\eg, image and species histogram) is provided in \cref{fig:stat_and_eval_protocal}a. To quantify class imbalance for a camera trap, we report two complementary metrics: (i) the fraction of images belonging to the two most frequent classes, and (ii) the number of images in the least frequent classes.

\begin{figure}[t]
    \centering
    \includegraphics[width=\linewidth]{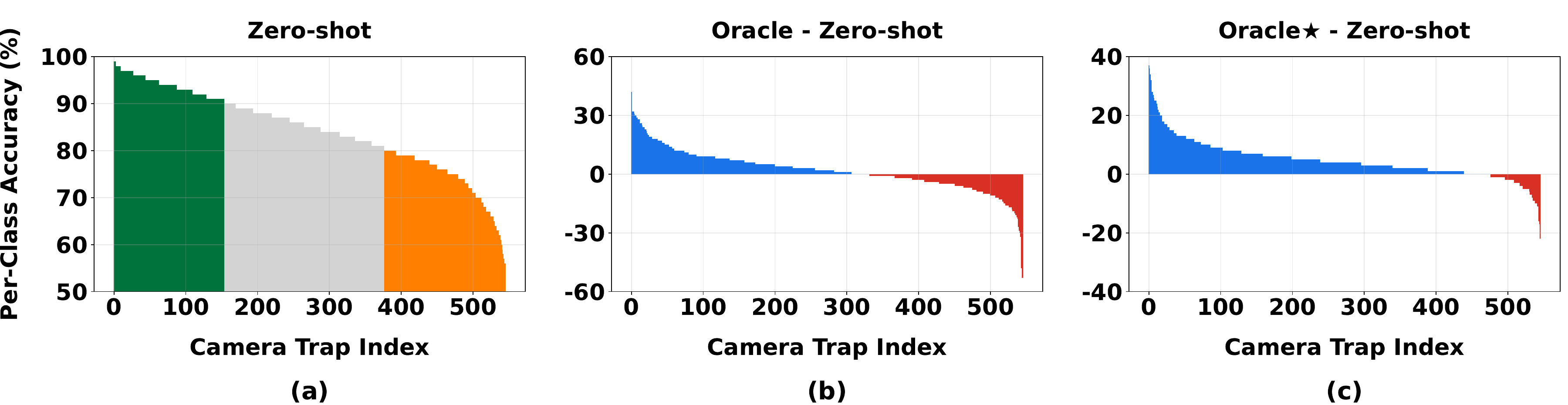}
    \vspace{-7mm}
    \caption{\small{{\textbf{(a).} Zero-shot accuracy across camera traps. } \textcolor{mossgreen}{\textbf{Green}}:  $>90\%$ accuracy; \textcolor{latexorange}{\textbf{orange}}: $<80\%$}. {\textbf{(b).} Accuracy difference between oracle naive fine-tuning and zero-shot.} \textcolor{posblue}{\textbf{Blue}}: oracle outperforms zero-shot; \textcolor{negred}{\textbf{red}}: zero-shot is better. {\textbf{(c).} Accuracy difference between oracle fine-tuning with our adaptation recipe (Oracle$\star$) and zero-shot.} } 
    \label{fig:zs_oracle_overall}
    \vspace{-0mm}
\end{figure}

\section{A Unified End-User-Centric Study}
\label{ss:unified_end_user}
Building on \textsc{StreamTrap}, we conduct the first unified \emph{end-user-centric} study to answer common questions faced by practitioners during deployment, and distill actionable guidelines for maintaining accurate models in dynamic environments.

\noindent\textbf{Setting \& Model.} For each camera trap, the model is initialized with the recently released BioCLIP 2~\cite{gu2025bioclip}, a biological vision foundation model trained on multimodal biological data and capable of classifying over $800$K species. To isolate the temporal dynamics we aim to study, we assume the candidate set of species expected at each camera trap is known a priori. This assumption aligns with real-world deployments, where practitioners typically have prior knowledge of the local fauna from historical observations or metadata. To ensure the broader generalizability of our findings, we provide additional experiments in~\cref{supp:relaxed_assumption} for scenarios where species are unknown beforehand.

\noindent\textbf{Evaluation Metric.} For each camera trap, we compute a model's average per-class accuracy within each interval (Balanced Accuracy), and then average these values across all intervals to obtain the overall performance.

\subsection{Is Adaptation Still Required with Foundation Models?}
\label{sec:adaptaion_required}

Before deployment, practitioners must decide whether a zero-shot biological foundation model is reliable enough for a new camera trap, or whether site-specific adaptation is still necessary. This question is especially important as recent biological foundation models~\cite{gu2025bioclip,stevens2024bioclip} already achieve state-of-the-art zero-shot accuracy across biodiversity datasets, including LILA BC subsets.

\noindent\textbf{Zero-shot baseline.} We begin by evaluating the zero-shot performance of BioCLIP 2 for all 546 camera traps in \textsc{StreamTrap}. To construct the zero-shot classifier head, we utilize each species' common name as the text label and apply OpenAI's standard prompt templates to generate the text embeddings. Formally, let $f_{\vtheta}$ denote the pre-trained vision encoder and $\vw_c$ the L2-normalized text embedding of class $c$. Given an image $\vx$, the predicted class is $\hat{y} = \argmax_c \vw_c^\top f_{\vtheta}(\vx)$. As illustrated in \cref{fig:zs_oracle_overall}a, we observe significant variability in zero-shot accuracy across 546 camera traps: BioCLIP 2 exceeds 90\% accuracy on 161 camera traps, highlighting its strong potential in wildlife monitoring, yet 162 camera traps fall in the 50--80\% range. This wide performance spread suggests that while zero-shot models provide a strong baseline, \textbf{there remains a persistent, critical need for site-specific adaptation.} We include zero-shot results for other foundation models in \cref{supp:zs_baseline} for reference.

%% file: sec/4_0_adapt_hard.tex
\subsection{Adaptation is Challenging in the Real World }
\label{sec:adapt_hard}

When off-the-shelf foundation models underperform at a camera trap, a natural next step is to \emph{adapt} them using labeled data collected at that site over time. The streaming evaluation protocol in \textsc{StreamTrap} mirrors this deployment lifecycle (\cref{sec: eval_proto}): for a given camera trap at interval $j$, the model is updated using cumulative labeled training data collected up to interval $j$, and is then evaluated on test data from the future interval $j{+}1$.


\noindent\textbf{Preliminary study.} We first examine supervised fine-tuning, arguably the most common adaptation strategy. We define the \textbf{accumulated model} as a baseline that naively fine-tunes on \emph{all} training data collected up to interval $j$. The model is initialized from BioCLIP 2, consistent with the zero-shot baseline above, and fine-tuned using standard cross-entropy loss with a cosine learning-rate scheduler and a base learning rate of $2.5 \times 10^{-5}$. Importantly, evaluation is always performed on unseen data from the future interval $j{+}1$.

To ensure the generalizability of our findings, we select 20 representative camera traps for this preliminary study, strategically sampled to span a wide spectrum of zero-shot accuracies. As shown in \cref{tab:20_cam}, naive accumulated fine-tuning \emph{consistently underperforms} the zero-shot baseline by a large margin. This result is particularly striking because the accumulated model is already an optimistic baseline: it uses all historical labeled data at each update step and imposes no explicit computation or storage constraint. Its failure therefore highlights the\textbf{ severe generalization challenge posed by realistic temporal adaptation in camera-trap deployments}, and motivates a deeper diagnosis of the factors driving adaptation failure. More experimental details and comprehensive results are provided in \cref{supp:add_analysis} and~\cref{supp:add_end_user}.

%% file: sec/4_1_investigation.tex
\input{tables_eccv/accum_20_new}
\subsection{Diagnosing the Drivers of Adaptation Challenges}
\label{sec:investigation}


We hypothesize that the severe adaptation difficulties may stem from two distinct but compounding factors: (1) the inter-interval distribution shifts driven by temporal non-stationarity; (2) the inherent difficulty of the camera trap data, even when temporal shift is reduced.

\noindent\textbf{Intrinsic difficulty even without temporal shift.} To disentangle these factors, we first consider a deliberately \emph{upper-bound} setting that removes the effect of inter-interval shift. Specifically, we train an \textbf{oracle} model using \emph{all} labeled training data across \emph{all} time intervals at a site, and evaluate it on held-out test data. This setting more closely resembles standard i.i.d. training than streaming adaptation, and one might therefore expect the oracle model to consistently outperform the zero-shot baseline. Surprisingly, this is not the case: on 214/546 camera traps ($\sim$40\%), the oracle model still underperforms the zero-shot model~(\cref{fig:zs_oracle_overall}b). This result indicates that temporal shift is not the sole source of difficulty; rather, the \emph{optimization and generalization challenges inherent to camera-trap data} are already substantial even when train and test distributions are more aligned.

A key driver of this difficulty is the passive nature of camera-trap data collection. Because camera traps must ``wait'' for animals to appear, many species are observed only rarely within a given site or time period, leading to severe class imbalance. As shown in \cref{fig:stat_and_eval_protocal}a, the two most frequent classes account for an average of $\sim$71\% of all images at a camera trap. In contrast, the two least frequent classes contain only $\sim$49 images on average, compared to $\sim$4,500 images for the two majority classes. Attempting to fine-tune a foundation model on such severely imbalanced, low-shot data inevitably introduces bias, amplifies memorization, and ultimately harms generalization.

\begin{figure}[t]
  \centering
  \includegraphics[width=\linewidth]{\detokenize{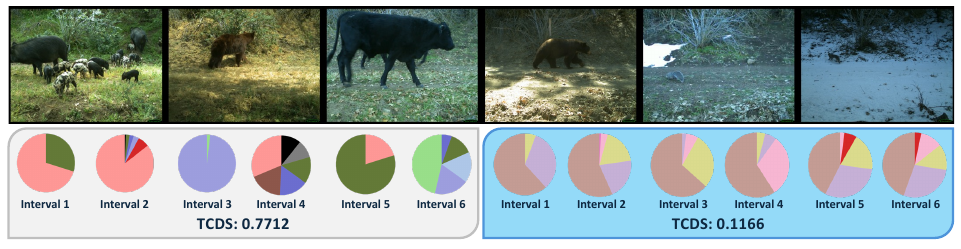}}
  \vspace{-5mm}
  \caption{\textbf{Illustration of temporal shifts.}
  (Top) Images from a camera trap showing shifting context (vegetation, lighting, weather) over time.
  (Bottom) TCDS comparison of two camera traps (pie chart: class frequency per interval): the left camera exhibits much higher temporal shifts (\textbf{TCDS=0.7712}) than the right one (\textbf{TCDS=0.1166}). }
  \label{dist_shift_demo}
  \vspace{-0mm}
\end{figure}
\noindent\textbf{Temporal shift is pervasive.}  The inherent non-stationarity of natural ecosystems causes both the background context (vegetation, illumination, weather) and the animal population distributions to undergo profound shifts between consecutive intervals, as illustrated in \cref{dist_shift_demo} (Top). To formally quantify the \textit{animal class distribution shift} between consecutive intervals, we propose the Temporal Class Distribution Shift (TCDS) metric for a specific camera trap. For class $c$ at interval $j$, let the normalized class frequency be $p_j^{c} = \frac{n_j^{c}}{\sum_{k} n_j^{k}}$, where $n_j^{c}$ is the sample count of class $c$ in interval $j$. We can define: 

\vspace{-0.2\baselineskip}
\begin{equation*}
    \begin{aligned}
        \scriptsize
        {\normalsize\textrm{TCDS}}\coloneqq
        \overbrace{\frac{1}{n-1}\sum_{j=1}^{n-1}}^{{\tiny\textit{\shortstack{\strut Average over\\[-2pt]\strut all intervals}}}}
        \!\!\!\!\!\!\!\!\underbrace{\sum_c\left|p_j^{c}-p_{j+1}^{c}\right|}_{{\tiny\textit{\shortstack{\strut Class distribution\\[-2pt]\strut shift between intervals}}}}
    \end{aligned}
\end{equation*}
\vspace{-0.2\baselineskip}




TCDS captures the average magnitude of temporal shift in class distributions across all consecutive intervals for a camera trap. As illustrated in \cref{dist_shift_demo} (Bottom), this metric quantifies class distribution shifts across camera traps, where higher TCDS values correspond to stronger temporal shifts. Unfortunately, a substantial fraction of the camera traps exhibit exceptionally high TCDS values (\cref{fig:stat_and_eval_protocal}a). Thus, a model that achieves high accuracy on the current interval can't guarantee to remain strong when confronted with the drastically shifted data distributions of future intervals.

\noindent\textbf{Compounding effects.} Severe data imbalance and temporal distribution shift are entangled in practice. Together, they make streaming adaptation substantially more challenging than standard in-domain fine-tuning: models must learn from scarce and imbalanced observations while also generalizing to future intervals whose species distributions may differ significantly from the past. This diagnosis motivates the need for robust adaptation recipes that help end-users maintain reliable model performance in dynamic deployment environments.

%% file: tables_eccv/accum_20_new.tex
\begin{table*}[t]
\centering
\scriptsize
\setlength{\tabcolsep}{2pt}
\renewcommand{\arraystretch}{1.5}
\setlength{\aboverulesep}{0pt}
\setlength{\belowrulesep}{0pt}

\resizebox{\textwidth}{!}{
\begin{tabular}{@{\hspace{2pt}}@{\extracolsep{\fill}} l w g w g w g w g w g w g w g w g w g w g w @{}}
\toprule
& {C1} & {C2} & {C3} & {C4} & {C5} & {C6} & {C7} & {C8} & {C9} & {C10} & {C11} & {C12} & {C13} & {C14} & {C15} & {C16} & {C17} & {C18} & {C19} & {C20} & Avg \\
\midrule
ZS & 97.7 & 92.3 & 90.9 & 89.1 & 86.1 & 84.3 & 81.5 & 81.2 & 78.3 & 77.1 & 75.8 & 73.8 & 70.1 & 69.8 & 67.1 & 66.0 & 62.2 & 61.5 & 56.3 & 55.8 & 75.8 \\
\midrule
Accum & 73.6 & 77.1 & 50.0 & 71.7 & 46.7 & 33.3 & 56.7 & 73.0 & 45.0 & 25.0 & 57.0 & 50.0 & 68.2 & 60.8 & 66.7 & 11.4 & 61.7 & 58.0 & 33.3 & 45.7 & 53.2 \\
$\Delta$ & \color{negred} 24.1 & \color{negred} 15.2 & \color{negred} 40.9 & \color{negred} 17.4 & \color{negred} 39.4 & \color{negred} 51.0 & \color{negred} 24.8 & \color{negred} 8.2 & \color{negred} 33.3 & \color{negred} 52.1 & \color{negred} 18.8 & \color{negred} 23.8 & \color{negred} 1.9 & \color{negred} 9.0 & \color{negred} 0.4 & \color{negred} 54.6 & \color{negred} 0.5 & \color{negred} 3.5 & \color{negred} 23.0 & \color{negred} 10.1 & \multicolumn{1}{c}{--} \\
\midrule
Accum$\star$ & 98.7 & 89.9 & 89.9 & 90.8 & 79.4 & 79.2 & 83.4 & 88.3 & 81.0 & 78.7 & 76.9 & 70.1 & 79.4 & 75.4 & 86.9 & 76.6 & 88.9 & 71.3 & 76.7 & 82.6 & 82.2 \\
$\Delta$ & \color{posblue} 1.0 & \color{negred} 2.4 & \color{negred} 1.0 & \color{posblue} 1.7 & \color{negred} 6.7 & \color{negred} 5.1 & \color{posblue} 1.9 & \color{posblue} 7.1 & \color{posblue} 2.7 & \color{posblue} 1.6 & \color{posblue} 1.1 & \color{negred} 3.7 & \color{posblue} 9.3 & \color{posblue} 5.6 & \color{posblue} 19.8 & \color{posblue} 10.6 & \color{posblue} 26.7 & \color{posblue} 9.8 & \color{posblue} 20.4 & \color{posblue} 26.8 & \multicolumn{1}{c}{--}\\
\midrule
{\color{gray} Oracle$\star$} 
& \color{gray} 98.7 & \color{gray} 96.4 & \color{gray} 94.5 & \color{gray} 96.6 & \color{gray} 87.8 & \color{gray} 91.5 & \color{gray} 91.0 & \color{gray} 91.0 & \color{gray} 87.3 & \color{gray} 82.0 & \color{gray} 84.3 & \color{gray} 88.5 & \color{gray} 91.0 & \color{gray} 86.6 & \color{gray} 91.0 & \color{gray} 83.9 & \color{gray} 93.8 & \color{gray} 89.6 & \color{gray} 89.9 & \color{gray} 93.4 & \color{gray} 90.4 \\
$\Delta$ & \color{posblue} 1.0 & \color{posblue} 4.1 & \color{posblue} 3.6 & \color{posblue} 7.5 & \color{posblue} 1.7 & \color{posblue} 7.2 & \color{posblue} 9.5 & \color{posblue} 9.8 & \color{posblue} 9.0 & \color{posblue} 4.9 & \color{posblue} 8.5 & \color{posblue} 14.7 & \color{posblue} 20.9 & \color{posblue} 16.8 & \color{posblue} 23.9 & \color{posblue} 17.9 & \color{posblue} 31.6 & \color{posblue} 28.1 & \color{posblue} 33.6 & \color{posblue} 37.6 & \multicolumn{1}{c}{--} \\
\bottomrule
\end{tabular}
}
\vspace{1mm}
\caption{\small {Accuracy difference ($\Delta$) between zero-shot (ZS) and three fine-tuning methods: naive accumulated fine-tuning (Accum), accumulated fine-tuning with our adaptation recipe (Accum$\star$), fine-tuned on all training data at a site with our recipe (Oracle$\star$). \textcolor{posblue}{\textbf{Blue}} (\textcolor{negred}{\textbf{red}}) denotes improvement (degradation) relative to ZS.}}
\label{tab:20_cam}
\vspace{-0mm}
\end{table*}

%% file: sec/5_accum.tex
\subsection{Practical Recipes for Robust Adaptation}
\label{sec:recipe}
We next investigate practical techniques for addressing the adaptation challenges identified above. Our goal is to establish an effective ``first-to-try'' recipe that end-users can apply when adapting foundation models to camera-trap deployments.

\subsubsection{Improving Generalization}
We first focus on strategies to mitigate the severe generalization degradation caused by extreme class imbalance and low-shot data. Motivated by prior literature on long-tailed recognition and efficient adaptation~\cite{zhang2023deep,mai2025lessons}, we investigate two primary directions:

\noindent\textbf{Class-imbalanced loss.} Standard cross-entropy fine-tuning on severely imbalanced data inherently biases the model toward majority classes, yielding poor generalization across classes~\cite{cui2019class,ye2020identifying,ye2021procrustean}. Class-imbalanced losses offer a simple and effective mitigation. We explore several established methods, including Balanced Softmax (BSM)~\cite{ren2020balanced}, Class-Balanced Focal Loss (CB-Focal)~\cite{cui2019class}, and Class-Dependent Temperatures (CDT)~\cite{ye2020identifying}. As detailed in our ablation studies (\cref{supp:ablation}), we find that BSM delivers consistent improvements across camera traps without requiring hyperparameter tuning,  whereas CB-Focal and CDT require more hyperparameter tuning, which is often infeasible for camera trap deployments. We therefore adopt BSM as our default imbalance-aware loss: $\mathcal{L}_{\text{BSM}}(\vx, y) =-\log\left(\frac{n_y \exp(\eta_y(\vx))}{\sum_{c} n_c \exp(\eta_c(\vx))}\right)$
where $\eta_c(\vx) = \vw_c^\top f_{\vtheta}(\vx)$ denotes the logit for class $c$, and $n_c$ is the number of training instances for class $c$.

\noindent\textbf{Parameter-Efficient Fine-Tuning (PEFT).} Fully fine-tuning on imperfect data often risks corrupting the generalized representations of foundation models~\cite{mai2024fine, zhang2025otvp}. Unlike full fine-tuning, which updates all network parameters, PEFT modifies only a minimal subset of weights, thereby preserving pre-trained knowledge~\cite{mai2025lessons, zhang2026revisiting}.  We evaluate three representative PEFT methods: Low-Rank Adaptation (LoRA)~\cite{hu2022lora}, Visual Prompt Tuning (VPT)~\cite{vpt, tu2023visual}, and Adapter~\cite{houlsbyadapter}. Our ablations (\cref{supp:ablation}) demonstrate that LoRA, which introduces trainable low-rank matrices into the attention projection layers, achieves superior performance without heavy hyperparameter tuning. Consequently, we adopt LoRA as our default fine-tuning mechanism.

\noindent\textbf{Synergy between imbalance loss and PEFT.} To rigorously assess the isolated and combined impacts of these two directions, we conduct an ablation study on a representative subset of 20 camera traps with the oracle setting (details in \cref{supp:ablation}). We observe that applying either the imbalance loss (BSM) or PEFT (LoRA) alone yields only marginal improvements (less than 1\%) over naive full fine-tuning. However, combining {BSM + LoRA} yields a substantial performance boost, strongly suggesting their compatibility and complementarity. Motivated by this result, we consider the \textbf{BSM + LoRA} as our recommended adaptation recipe for camera traps (denoted hereafter as $\star$). Applying this recipe to the oracle models (Oracle$\star$) across all 546 camera traps in the \textsc{StreamTrap} benchmark enables 474 sites (compared to only 332 for the naive oracle) to successfully outperform the zero-shot baseline (\cref{fig:zs_oracle_overall}c). While there remain 72 camera traps where this recipe falls short, we provide a detailed qualitative and quantitative analysis of these specific failure cases in \cref{supp:remaining_oracle_failure}.

\subsubsection{Mitigating Temporal Shifts}

Our proposed adaptation recipe ($\star$) successfully transfers its effectiveness from the idealized oracle setting to the accumulated model evaluated under the realistic streaming protocol. As shown in \cref{tab:20_cam}, the accumulated model equipped with our recipe (denoted as Accum$\star$) consistently outperforms the zero-shot baseline. However, a non-negligible performance gap persists between Accum$\star$ and corresponding upper bounds (Oracle$\star$) on most camera traps. This discrepancy is primarily driven by inter-interval distribution shifts, which inherently complicate streaming evaluations. Mitigating such temporal shifts is notoriously difficult due to the unpredictability of future data distributions, a challenge that remains largely underexplored in the vision community. To bridge this gap, we evaluate various approaches and identify three promising post-processing techniques that can potentially improve Accum$\star$.

\noindent\textbf{Post-hoc logit calibration.} 
When some classes are absent or extremely low-shot during fine-tuning, the updated model tends to heavily suppress their predicted logits relative to the majority classes, even when the foundation model originally represented them well~\cite{tu2023holistic,mai2024fine}. To rectify this bias, we adopt a post-hoc logit calibration~\cite{mai2024fine} to boost the logits of absent or minority classes to a magnitude comparable to the majority classes by introducing a calibration factor $\gamma$:
$\hat{y} = \argmax_c \vw_c^\top f_{\vtheta}(\vx) + \gamma \cdot \mathbf{1}[c \in \mathcal{A}]$, where $\mathcal{A}$ denotes absent/minority classes in the current interval.

\noindent\textbf{Weight interpolation.} While fine-tuning enables necessary task-specific adaptation, it risks forgetting of the foundation model's broad generalization capabilities. To strike a balance between generalizable pre-trained knowledge and site-specific features, we apply weight interpolation (WiSE)~\cite{wortsman2022robust}. Let $\vtheta'$ be the fine-tuned weights and $\vtheta$ the pre-trained weights. The interpolated weights are given by $ \vtheta(\alpha) = \alpha \vtheta + (1 - \alpha) \vtheta'$ where $\alpha \in [0, 1]$ controls the interpolation ratio.

\noindent\textbf{Interval model selection.} The most recently adapted model is not always the best model for the next interval. Due to recurring seasonal patterns or abrupt ecological changes, future interval $j{+}1$ may resemble an earlier interval more than the immediately preceding one. To assess the potential benefit of choosing among historical models, we conduct a diagnostic upper-bound study: for each future interval $j{+}1$, we select the historical or current model that achieves the highest accuracy on that interval's test split. This procedure is not directly deployable because it uses future labels, but it quantifies the possible gain from better model-selection mechanisms and motivates future work on practical selection criteria under streaming deployment.

\begin{figure}[t]
\centering
\includegraphics[width=1.0\columnwidth]{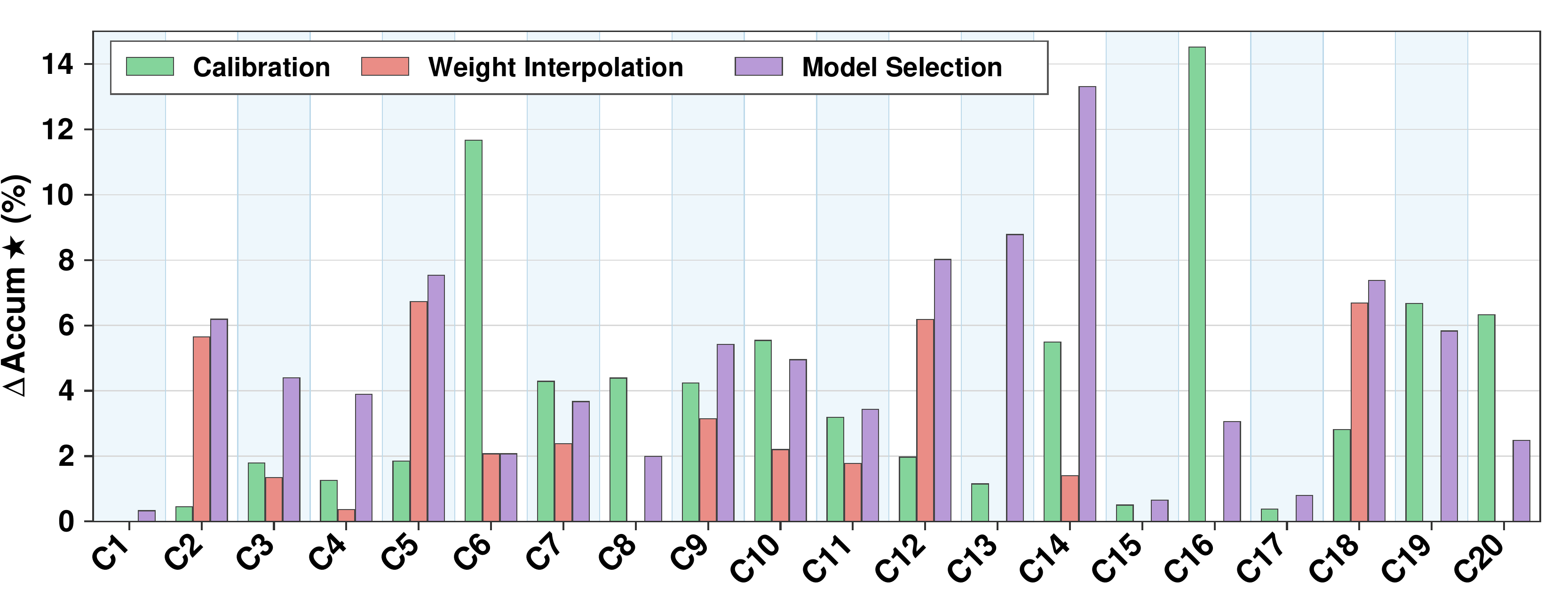}
\vspace{-5mm}
\caption{\small Performance gain ($\Delta_{\text{Accum}\star}$) of calibration, weight interpolation, and model selection over the Accum$^\star$ baseline. They are applied on top of Accum$^\star$. More results can be found in~\cref{supp:more_post_processing}.}
\label{fig:delta_plot}
\vspace{-0mm}
\end{figure}



\noindent\textbf{Promises and limits of post-processing.} \cref{fig:delta_plot} illustrates the relative gains of each post-adaptation strategy over Accum$\star$. Notably, for camera traps where Accum$\star$ initially underperforms the zero-shot baseline, at least one of these strategies can recover the performance drop. Moreover, these techniques can effectively narrow the gap between the Oracle$\star$ and Accum$\star$, underscoring their tremendous potential for mitigating temporal shift. However, we emphasize that these results are obtained under an optimistic upper-bound setting, where we select performance-maximizing hyperparameters to reveal the potential of each method. In real deployments, selecting such hyperparameters without access to future data remains an open challenge. We therefore view this as an important research opportunity and encourage future work to resolve the degradation caused by temporal shifts.

\begin{tcolorbox}
[colback=gray!5,colframe=gray!60, title=\text{\small Recommended Adaptation Recipes}]
\small
We recommend combining \textbf{LoRA} (PEFT) with \textbf{BSM} (imbalance loss) as a practical first-to-try adaptation strategy, augmented by post-processing techniques, such as \textbf{logit calibration}, \textbf{weight interpolation}, or \textbf{interval model selection}, to mitigate inter-interval distribution shift.
\end{tcolorbox}

%% file: sec/6_end_user.tex
\section{More Practical Considerations}

\begin{figure}[tbh]
\centering
\includegraphics[width=\linewidth]{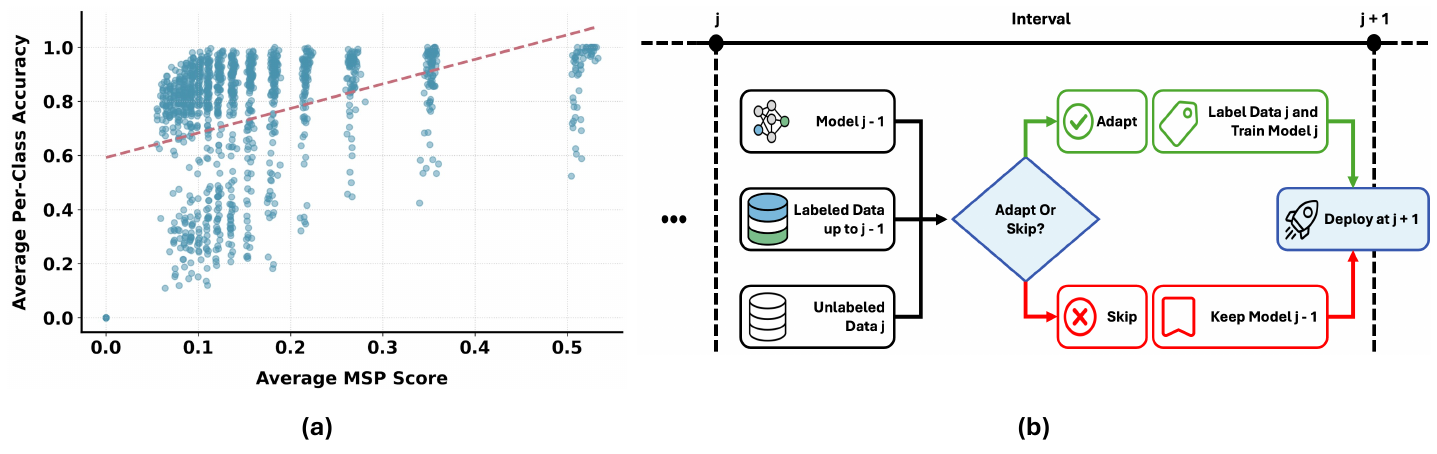}
\caption{\small \textbf{(a).} Non-OOD scores (MSP) moderately correlate with zero-shot accuracy. \textbf{(b).} The Adapt-or-Skip problem: should we adapt at this interval or not?.}
\label{fig:OOD-MS}
\end{figure}

Accurate real-world deployment involves more than selecting an adaptation algorithm. Practitioners must also decide when a foundation model is sufficient, whether continual adaptation is necessary, and when adaptation should be triggered. Despite their importance for deployment, these questions remain largely underexplored in the vision community. In this section, we provide preliminary analyses with simple baselines and highlight directions for future research.


\subsubsection{When is the zero-shot model sufficient?} In real-world deployments, practitioners need to estimate whether a foundation model’s zero-shot performance is sufficient for a new site. If yes, the model can be deployed directly without data collection and adaptation. However, this decision must typically be made before labeled data from the new site are available, making zero-shot accuracy hard to assess in advance. Despite its practical importance, this problem remains largely underexplored in the vision community. As a simple baseline, we assess if out-of-distribution (OOD) signals can provide an indication of zero-shot accuracy~\cite{yang2024generalized}. The intuition is that if images from a new camera trap are closer to the training distribution, the model may produce higher-confidence predictions and achieve better zero-shot accuracy. Specifically, we adopt the Maximum Softmax Probability (MSP)~\cite{hendrycks2016baseline}, assigning each test sample a non-OOD score:
$s_{\text{MSP}}(\vx) = \max_c 
\frac{\exp(\eta_c(\vx)/\tau)}{\sum_{c'} \exp(\eta_{c'}(\vx)/\tau)}$
where $\eta_c(\vx) = \vw_c^\top f_{\vtheta}(\vx)$ denotes the logit for class $c$, and $\tau$ is a temperature parameter. We plot the average non-OOD score for each camera trap against the zero-shot accuracy in \cref{fig:OOD-MS}a. While a moderate correlation can be observed, many camera traps deviate from this pattern, indicating that confidence signals alone are insufficient to reliably predict zero-shot performance. This highlights the difficulty of estimating zero-shot sufficiency prior to deployment, suggesting it as an open problem for future work.





\subsubsection{Do we need continual adaptation?} 


Another practical question raised by end-users is whether continual adaptation is necessary after a model has already been updated over several initial intervals. Once the model has learned site-specific characteristics, can it generalize to all future data without further fine-tuning? To investigate this question, we conduct a controlled study in which adaptation is deliberately halted after a fixed number of intervals. Specifically, for camera traps with long deployment durations spanning more than one year, we freeze the accumulated model at a specific interval and evaluate it on all remaining future intervals. As shown in \cref{tab:ckp2_4_last_split}, model accuracy generally improves as the model is adapted on more intervals. This trend indicates that continually incorporating additional site-specific data can provide cumulative performance gains. At the same time, the benefit of each update is not uniform: some intervals provide substantially richer adaptation signals than others, while others offer limited marginal improvement. This observation motivates our next question: can we determine \emph{when} an adaptation step is actually necessary?

\input{tables_eccv/continue_small_no_wrap}

\subsubsection{When should we adapt?}
\label{ss:when_adapt}
Adaptation incurs labeling and computational costs, making it important to determine when an update is actually beneficial. However, this decision must be made before observing future performance. At every interval, the end-users will face an \textbf{Adapt-or-Skip} problem as illustrated in \cref{fig:OOD-MS}b: should we adapt at this interval or not?  Despite its practical importance, this remains largely underexplored. At interval \(j\), the user observes the deployed model \(f_{j-1}\), prior labeled training data \(\mathcal{D}_{1:j-1}^{\mathrm{train}}\), and current unlabeled samples \(\mathcal{X}_j\), and must decide whether to \textsc{Adapt} or \textsc{Skip}. If adapting, the user labels \(\mathcal{X}_j\) and trains $f_j$ with $\mathcal{D}_{1:j}^{\mathrm{train}}$; if skipping, they keep deploying \(f_{j-1}\). Retrospectively, the ground-truth decision is \(a_j^\star=\textsc{Adapt}\) iff \(\operatorname{Acc}(f_j,\mathcal{D}_{j+1}^{\mathrm{test}})>\operatorname{Acc}(f_{j-1},\mathcal{D}_{j+1}^{\mathrm{test}})\), and \(a_j^\star=\textsc{Skip}\) otherwise. The challenge is to predict \(a_j^\star\) using only information available at interval \(j\), before observing \(\mathcal{D}_{j+1}^{\mathrm{test}}\). We construct a dataset where half of the intervals favor adapting and half favor skipping. We evaluate three simple baseline methods: {random}, {MSP-based}, and {CLIP feature-based}. However, both methods perform similarly to random guessing, highlighting the difficulty of the problem. For reference, we report an upper-bound that always selects the correct decision, which outperforms the three baselines by {11.34\%}, suggesting substantial room for future research. More details are provided in \cref{supp:when_adapt}.

%% file: tables_eccv/continue_small_no_wrap.tex
\begin{table}[tbh]
\centering
\scriptsize
\setlength{\tabcolsep}{8pt}
\renewcommand{\arraystretch}{1.0}
\begin{tabular}{lccccc}
    \toprule
    \textbf{Stage} & $0\%$ & $25\%$ & $50\%$ & $75\%$ & $100\%$ \\
    \midrule
    \textbf{Acc.} & 76.2 & 73.1 & 77.8 & 82.7 & 84.9 \\
    \bottomrule
\end{tabular}
\vspace{1mm}
\caption{\small Accum$\star$ is frozen at a specific interval (\eg, 25\% means frozen after adaptation with the first 25\% intervals) and evaluated on all future intervals. Averaged across 8 camera traps with long deployment durations spanning more than one year.}
\label{tab:ckp2_4_last_split}
\end{table}

%% file: sec/7_conclusion.tex
\section{Conclusion}


We present the first unified study of camera-trap species recognition over time, prioritizing reliable, long-term fixed-site deployment. Our benchmark, \textsc{StreamTrap}, uses a streaming evaluation protocol reflecting real-world data collection. Because naive foundation model fine-tuning fails under extreme data imbalance and temporal shifts, we provide robust adaptation recipes and analyze critical deployment challenges to guide practitioners and spur future research.

\section*{Acknowledgment}
This research is supported by grants from the National Science Foundation (ICICLE: OAC-2112606). We are grateful for the generous support from the Ohio Supercomputer Center.

%% file: sec/X_suppl_eccv.tex
\clearpage
\setcounter{page}{1}

\makeatletter
\@ifpackageloaded{hyperref}{%
  \renewcommand*{\theHsection}{appendix.\Alph{section}}%
  \renewcommand*{\theHsubsection}{appendix.\Alph{section}.\arabic{subsection}}%
  \renewcommand*{\theHsubsubsection}{appendix.\Alph{section}.\arabic{subsection}.\arabic{subsubsection}}%
}{}
\makeatother

\appendix
\crefname{appendix}{Appendix}{Appendices}
\Crefname{appendix}{Appendix}{Appendices}
\crefalias{section}{appendix}
\crefalias{subsection}{appendix}
\crefalias{subsubsection}{appendix}
\noindent{In this supplementary material, we provide additional details and experimental results that complement the main paper. Specifically, we include:}
\vspace{1mm}
\begin{itemize}
    \item \cref{supp:related_work}: Detailed Related Work
    \item \cref{supp:add_analysis}: Adaptation Recipe: Training Details and Ablations
    \item \cref{supp:add_end_user}: Additional Details and Results
    \item \cref{supp:benchmark}: \textsc{StreamTrap}: Construction Details
\end{itemize}
\vspace{-5mm}

\section{Detailed Related Work}
\label{supp:related_work}

\subsection{Camera Trap Data in Computer Vision}
Large-scale wildlife image collections captured through camera traps have emerged as invaluable resources for biodiversity monitoring, providing critical insights into species richness, occupancy patterns, and animal behaviors~\cite{trolliet2014use, boitani2016camera}. These automated imaging systems generate substantial volumes of visual data, typically requiring extensive off-site manual analysis. To alleviate this bottleneck, deep neural networks have been increasingly adopted for automating tasks such as species detection and classification, positioning camera trap datasets as pivotal resources within the computer vision and machine learning communities~\cite{norouzzadeh2018automatically, yu2013automated}.

A significant research direction involves the generalization capabilities of models trained on camera trap data~\cite{koh2021wilds, mai2024fine}. Specifically, systems trained to detect and classify animals are frequently deployed in novel geographical locations without subsequent fine-tuning. To accurately reflect this practical constraint, the widely recognized iWildCam challenges~\cite{beery2019iwildcam, beery2021iwildcam} deliberately partition their datasets by camera location. This ensures no overlap between training and test camera locations, providing a robust evaluation of model generalization to unseen environments. Additionally, with the advent of multimodal foundation models, camera trap imagery has increasingly been integrated into multimodal analysis frameworks to enhance wildlife monitoring through richer contextual understanding~\cite{gabeff2024wildclip, fabian2023multimodal}. In parallel, CATALOG~\cite{santamaria2025catalog} demonstrates that combining multiple foundation models within a contrastive learning framework improves zero-shot camera trap recognition across Snapshot Serengeti~\cite{swanson2015serengeti} and Terra Incognita~\cite{beery2018recognition}. Recent work also emphasizes label efficiency and practitioner usability: Bothmann et al. propose an active-learning-based pipeline that couples object detection, careful hyperparameter tuning, and iterative query strategies to train accurate wildlife classifiers from relatively small, region-specific camera trap datasets~\cite{bothmann2023automated}. 

While existing literature largely frames camera trap analysis as a cross-domain generalization problem, this perspective diverges from the practical needs of end-users such as ecologists and conservation practitioners. Their primary concern is sustaining reliable species recognition at a \textbf{fixed deployment site over time}, where seasonal variations, habitat changes, and animal migration continuously reshape both animal appearances and backgrounds, causing model performance to degrade substantially over extended deployments. To address this critical gap, we conduct a \textbf{unified, end-user-centric study} that reframes evaluation around the actual deployment lifecycle, systematically characterizing adaptation challenges and deriving actionable guidelines for long-term camera trap deployment.

\subsection{Continual Learning}
Unlike conventional domain adaptation settings~\cite{farahani2021brief, zhang2025prime}, where data with distribution shifts are available all at once, continual learning deals with models that must adapt to non-stationary data streams over time~\cite{mai2022online, zhang2025dpcore, lee2025continual}. A key challenge is maintaining reliable performance as distributions shift, whether from new classes, changing backgrounds, or evolving environmental conditions~\cite{aljundi2019gradient, shim2021online}. However, most existing benchmarks artificially partition datasets without actual timestamps, creating evaluation scenarios that are either unrealistically simple or excessively challenging~\cite{zeroflow, erd}.

Recently, Velasco-Montero et al. demonstrated how continual learning could be embedded into smart camera traps for efficient deployment, but their focus was primarily on hardware and system design~\cite{velasco2024reliable}. Zhu et al. studied class-incremental learning for wildlife monitoring but did not account for the real temporal order in camera trap data~\cite{zhu2022class}. More broadly, conventional continual learning assumes limited access to past data due to privacy or storage constraints, an assumption that does not hold in ecological deployments where labeled camera trap data is a persistent, archivable resource. We therefore train on the full historical record at each interval to reflect this reality, naturally exposing temporal non-stationarity as the central challenge, and providing a real-world testbed grounded in genuine ecological dynamics for the broader continual learning community.

\subsection{Class-Imbalanced Learning}
\label{supp:related_class_imbalance}
The distribution of animal classes in camera trap datasets is often highly imbalanced. Models trained on such an imbalanced dataset are often heavily biased, performing exceptionally well on majority classes while underperforming significantly on minority classes~\cite{bevan2024deep, malik2021two}. However, models that achieve balanced performance across all classes are more valuable, especially considering that rare classes are often of particular ecological interest. Techniques designed to mitigate class imbalance in deep learning can be broadly categorized into two groups: data-level methods, which modify the data distribution directly by oversampling or undersampling, and algorithm-level methods, which adjust the learning process itself through specialized loss functions or other algorithmic modifications. For a comprehensive review of these approaches, we refer readers to recent surveys~\cite{zhang2023deep, rezvani2023broad}. In our study, we specifically explore simple yet highly effective algorithm-level methods, such as the Balanced Softmax loss~\cite{ren2020balanced}, to address class imbalance issues in camera trap datasets. 


\section{Adaptation Recipe: Training Details and Ablations}
\label{supp:add_analysis}

\subsection{Zero-shot Baseline}
\label{supp:zs_baseline}
We adopt BioCLIP~2~\cite{gu2025bioclip} as our zero-shot baseline, as it delivers the strongest zero-shot performance on all camera traps in \textsc{StreamTrap}. As shown in \cref{tab:zs_comparison}, BioCLIP~2 achieves an average zero-shot per-class accuracy of 84.3\%, substantially outperforming CLIP~\cite{openaiCLIP}. This gap reflects BioCLIP~2's pre-training on large-scale biological imagery, which enables it to better capture the fine-grained appearance variation inherent to camera trap data.

\input{tables_eccv/zs_comparison}

\subsection{Training Details}
We fine-tune BioCLIP~2 using the configuration summarized in \cref{tab:training_config}. We employ early stopping based on validation performance. For the oracle model, the validation set is constructed by randomly sampling two images per class from the training set. For the accumulated model at interval $j$, we use the test split of the preceding interval $j{-}1$ as the validation set.

\input{tables_eccv/train_details}

\subsection{Ablation of Imbalance Loss and PEFT}
\label{supp:ablation}
We ablate imbalance losses and PEFT methods under the oracle setting to identify the most effective combination. Since running all configurations across all 546 camera traps in \textsc{StreamTrap} is computationally prohibitive, we strategically subsample 20 camera traps that span the full range of accuracy gaps between the naively fine-tuned oracle and the zero-shot baseline. This ensures that our ablation findings generalize across both easy and challenging deployment scenarios.
\input{tables_eccv/oracle_loss_peft}

\noindent\textbf{Imbalance loss comparison.}
As shown in Fig.~2a in the main paper, severe class imbalance is a defining characteristic of camera traps in \textsc{StreamTrap}, where a few dominant species account for most observations while many others appear only sparsely. Training with standard cross-entropy biases the classifier toward majority classes, diminishing its ability to recognize minority species~(\cref{supp:related_class_imbalance}). To evaluate imbalance-aware training objectives, we compare three widely used methods: Balanced Softmax (BSM)~\cite{ren2020balanced}, Class-Balanced Focal Loss (CB-Focal;~$\beta{=}0.999$, $\gamma{=}0.5$)~\cite{cui2019class}, and Class-Dependent Temperatures (CDT;~$\gamma{=}0.3$)~\cite{ye2020identifying}. Notably, BSM requires no additional hyperparameters, whereas CB-Focal and CDT require tuning.

\noindent\textbf{PEFT comparison.}
Unlike full fine-tuning, which risks corrupting the generalized representations of the foundation model, PEFT methods update only a minimal subset of parameters to better preserve pre-trained knowledge. We compare three representative approaches: Low-Rank Adaptation (LoRA; $r{=}8$)~\cite{hu2022lora}, which introduces learnable low-rank matrices into all attention projection layers; Visual Prompt Tuning (VPT-Deep; 10 prompt tokens per layer)~\cite{vpt}, which steers the model by augmenting the input sequence with learnable tokens; and Adapter Tuning (bottleneck dimension${}=8$, $\alpha{=}1.0$)~\cite{houlsbyadapter}, which inserts lightweight bottleneck modules after both the attention and MLP sublayers. In all cases, only the newly introduced parameters are updated while the backbone remains frozen.

\noindent\textbf{From ablation to recipe.}
As shown in \cref{tab:loss_peft}, BSM yields the best average performance among the three imbalance losses, while CB-Focal and CDT exhibit performance degradation. For PEFT, LoRA consistently outperforms both VPT and adapter tuning, aligning with recent studies showing its stability in low-data regimes. We further examine the compatibility of the best-performing method from each direction. As shown in \cref{tab:loss_peft}, BSM and LoRA are not only compatible but complementary: their combination yields the strongest and most consistent performance, where LoRA contributes stable and efficient adaptation while BSM effectively mitigates the severe class imbalance inherent in camera-trap streams. We therefore adopt BSM and LoRA as our recommended adaptation recipe~($\star$).




\section{Additional Details and Results}
\label{supp:add_end_user}

\subsection{Evaluating Beyond Assumptions}
\label{supp:relaxed_assumption}
In the main paper (\cref{ss:Fair Data Processing Pipeline} and \cref{ss:unified_end_user}), we adopt two assumptions to maintain evaluation stability and mirror real-world deployment: a closed-set species vocabulary and the exclusion of rare species within each interval. To ensure the generalizability of our findings, we relax each assumption below. For both analyses, we evaluate ten camera traps sampled randomly.
\input{tables_eccv/openset_and_rare}

\noindent\textbf{Open-set classification.}
In \cref{ss:unified_end_user} of the main paper, we assume the candidate species at each camera trap are known a priori (closed-set), allowing us to isolate temporal dynamics. Here, we relax this assumption by constructing an open-set vocabulary from all 1,096 classes across \textsc{StreamTrap} and initializing models with this expanded class list (open-set). This setting is more representative of real-world deployments where practitioners may not have complete prior knowledge of which species are present at a given site. It also tests whether the adaptation recipe remains effective when the model must discriminate among a substantially larger set of candidate species.

\noindent\textbf{Including rare species.}
As described in \cref{ss:Fair Data Processing Pipeline} of the main paper, rare species (fewer than 10 samples per interval) are excluded from evaluation to prevent severe class imbalance. However, rare species are often of greatest ecological interest, as they may include threatened or endangered populations that require urgent conservation attention. To reflect this, we include them exclusively in the test set, as they lack sufficient data for a train/test split. This setup evaluates whether adapted models can still recognize species that were never or barely seen during training.

\noindent\textbf{Results and analysis.}
The results in \cref{tab:openset_and_rare} indicate that all models perform worse under the open-set setting compared to the closed-set configuration, as the enlarged label space introduces additional confusion among visually similar species. Including rare species also degrades all models, but the oracle$\star$ model suffers the largest drop, falling below both the zero-shot baseline and the accum$\star$ model. We hypothesize that the oracle$\star$ model over-adapts to dominant common classes across all pooled data, eroding the foundation model's ability to recognize inherently rare species. The accum$\star$ model, by adapting incrementally, better preserves general representations and remains competitive with the zero-shot baseline.



\subsection{Remaining Failure Cases of Oracle$\star$}
\label{supp:remaining_oracle_failure}

\begin{figure}[tbh]
\small
\centering
\noindent\includegraphics[width=1\linewidth]{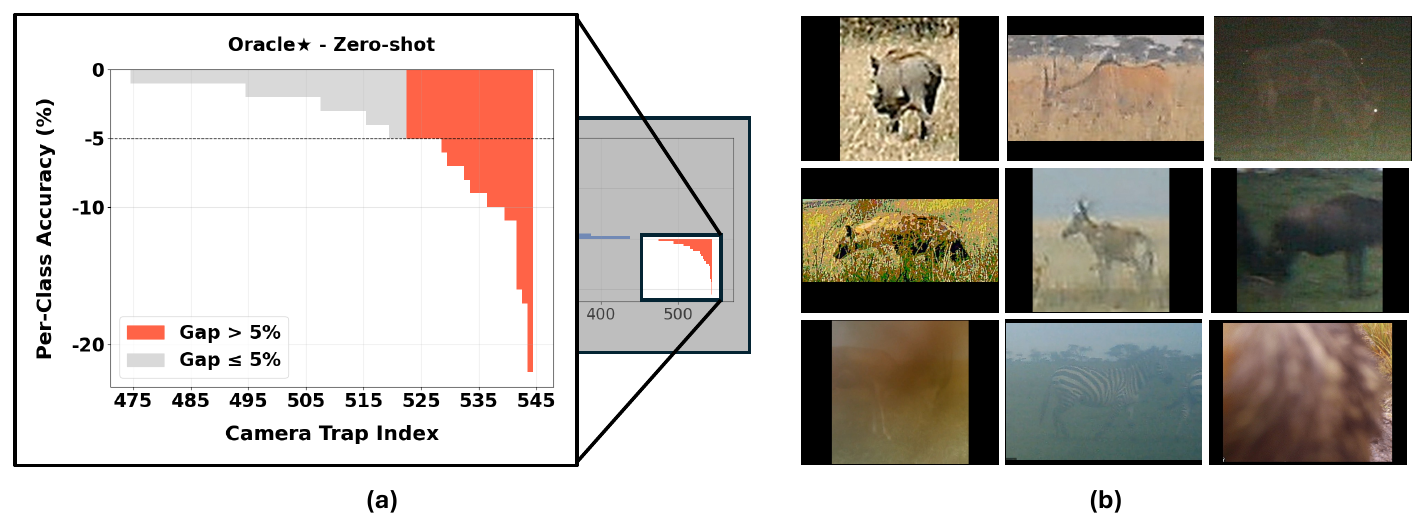}
\caption{\small \textbf{(a.)} Detailed accuracy difference between oracle fine-tuning with our adaptation recipe (Oracle$\star$) and zero-shot (\cref{fig:zs_oracle_overall} in the main paper). \textbf{(b.)} Qualitative examples of camera traps with severely degraded visual quality (e.g., blur, exposure artifacts, and sensor noise).}
\label{fig:oracle_recipe_failure}
\vspace{-5mm}
\end{figure}

While our proposed approach generally improves the performance of the oracle models compared to the zero-shot baseline, 72 camera traps still underperform it (\cref{fig:zs_oracle_overall} in the main paper). As illustrated in \cref{fig:oracle_recipe_failure}a, 48 of these 72 show a gap of at most 5\%. However, the remaining 22 camera traps exhibit a more substantial degradation exceeding 5\%. To better understand these limitations, we conduct a detailed analysis of these 22 camera traps, separating minor fluctuations from more systematic failure modes.

\noindent\textbf{Poor image quality (5--10\%).}
For the 16 camera traps with a moderate performance gap (between 5 and 10\%), the degradation is primarily driven by poor image quality. To accurately mirror real-world deployment conditions, we intentionally retain low-quality images in which animals are present but difficult to discern. As illustrated in \cref{fig:oracle_recipe_failure}b, the animal is visible upon close inspection, yet the degraded image quality introduces noise that can mislead the fine-tuned model---whereas the zero-shot baseline, not having been exposed to these noisy training signals, remains unaffected. Filtering such low-quality images could improve overall model performance, but we consider this outside the current scope and leave it as a direction for future work.

\noindent\textbf{Train--test distribution shift ({>}10\%).}
For the 6 camera traps exhibiting a performance drop of more than 10\%, the degradation consistently traces back to a severe distribution shift between the training and test splits. As described in \cref{sec:add_details_data_process}, \textsc{StreamTrap} groups images from the same burst before splitting to reduce near-duplicate leakage across splits. While this makes the evaluation more realistic, it can also produce splits with markedly different visual conditions for some camera traps. For example, the training split may contain mostly daytime images, whereas the test split contains mostly nighttime images. Such a distribution mismatch appears to be the primary factor behind the largest failure cases, while the zero-shot baseline retains general knowledge that transfers across both conditions.



\subsection{Post-Processing: Extended Evaluation}
\label{supp:more_post_processing}
\begin{figure}[tbh]
\centering
\vspace{-3mm}
\includegraphics[width=\textwidth]{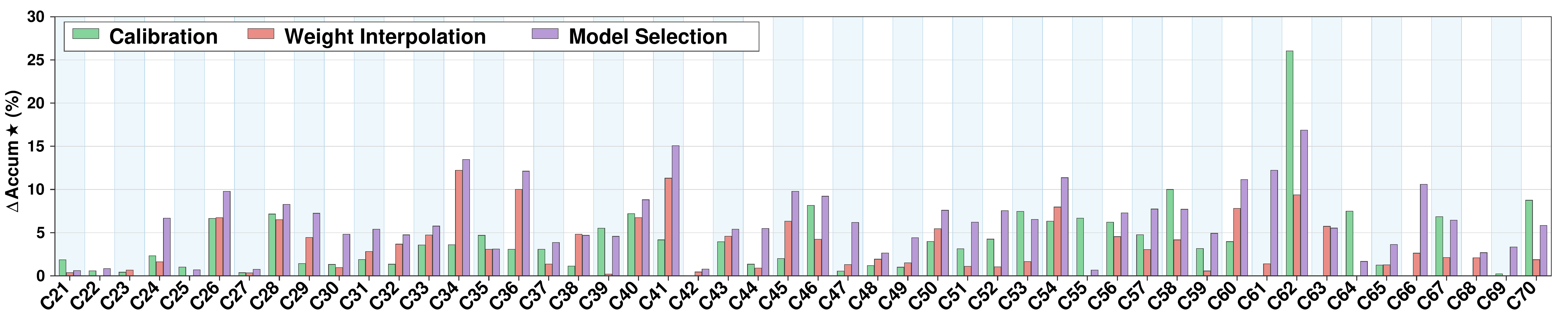}
\caption{Performance gain ($\Delta{\text{accum}\star}$) of calibration, weight interpolation, and model selection over the accum$^\star$ baseline across 50 cameras. They are applied on top of accum$^\star$.}
\label{fig:postproc_plot}
\vspace{-15mm}
\end{figure}

To complement the post-processing analysis, we evaluate all three post-processing methods on 50 additional camera traps where oracle$\star$ outperforms the zero-shot baseline, sampled to span a diverse range of accuracy gaps between the two models. As shown in \cref{fig:postproc_plot} and \cref{tab:postproc}, all methods yield promising improvements over the accum$\star$ baseline, and selecting the best-performing post-processing method per camera trap further narrows the gap with oracle$\star$. This suggests that with principled hyperparameter selection, post-processing can substantially boost adapted models. Developing automatic hyperparameter selection mechanisms that do not require access to future data remains a promising direction for future work.

\input{tables_eccv/appendix_post_process}

\subsection{Details of When Should We Adapt}
\label{supp:when_adapt}
As illustrated in \cref{fig:OOD-MS}b of the main paper, practitioners must decide whether to adapt the model when a new interval of data arrives. We construct an adapt-or-skip benchmark on top of \textsc{StreamTrap} to evaluate this. Throughout this section, we denote by $\mathrm{model}_j$ the model trained on cumulative labeled data from intervals $1,\dots,j$.

\noindent\textbf{Experimental setup.}
Following the streaming protocol, at interval $j$, the learner has access to $\mathrm{model}_{j-1}$ and the unlabeled images from the current interval $j$. The learner must decide whether to:
\begin{itemize}[label=$\bullet$]
\item \textbf{Adapt}: label the interval $j$ data and train $\mathrm{model}_j$.
\item \textbf{Skip}: continue using $\mathrm{model}_{j-1}$.
\end{itemize}
This decision must be made \emph{before observing the test data from interval \(j+1\)}. The chosen model (either \(\mathrm{model}_{j-1}\) or \(\mathrm{model}_j\)) is then evaluated on the test split of interval \(j+1\). Each interval therefore, produces one adapt-or-skip decision instance. To construct the benchmark, we uniformly sample 80 camera traps from those where the oracle$\star$ performance exceeds the zero-shot baseline, and subsample decision instances to obtain a balanced dataset with equal numbers of Adapt and Skip cases. We evaluate three simple decision heuristics:

\noindent\textbf{Random choice.} Selects Adapt or Skip uniformly at random.

\noindent\textbf{MSP-based confidence method.}
We compute the mean Maximum Softmax Probability (MSP) of \(\mathrm{model}_j\) on the unlabeled images of interval \(j\). The intuition is that model confidence can provide a coarse signal of distribution shift. Adaptation is triggered when the confidence falls below a threshold \(\tau_{\mathrm{MSP}}(j)\).

\begin{figure}[t]
    \centering
    \begin{subfigure}{0.48\linewidth}
        \centering
        \includegraphics[width=\linewidth]{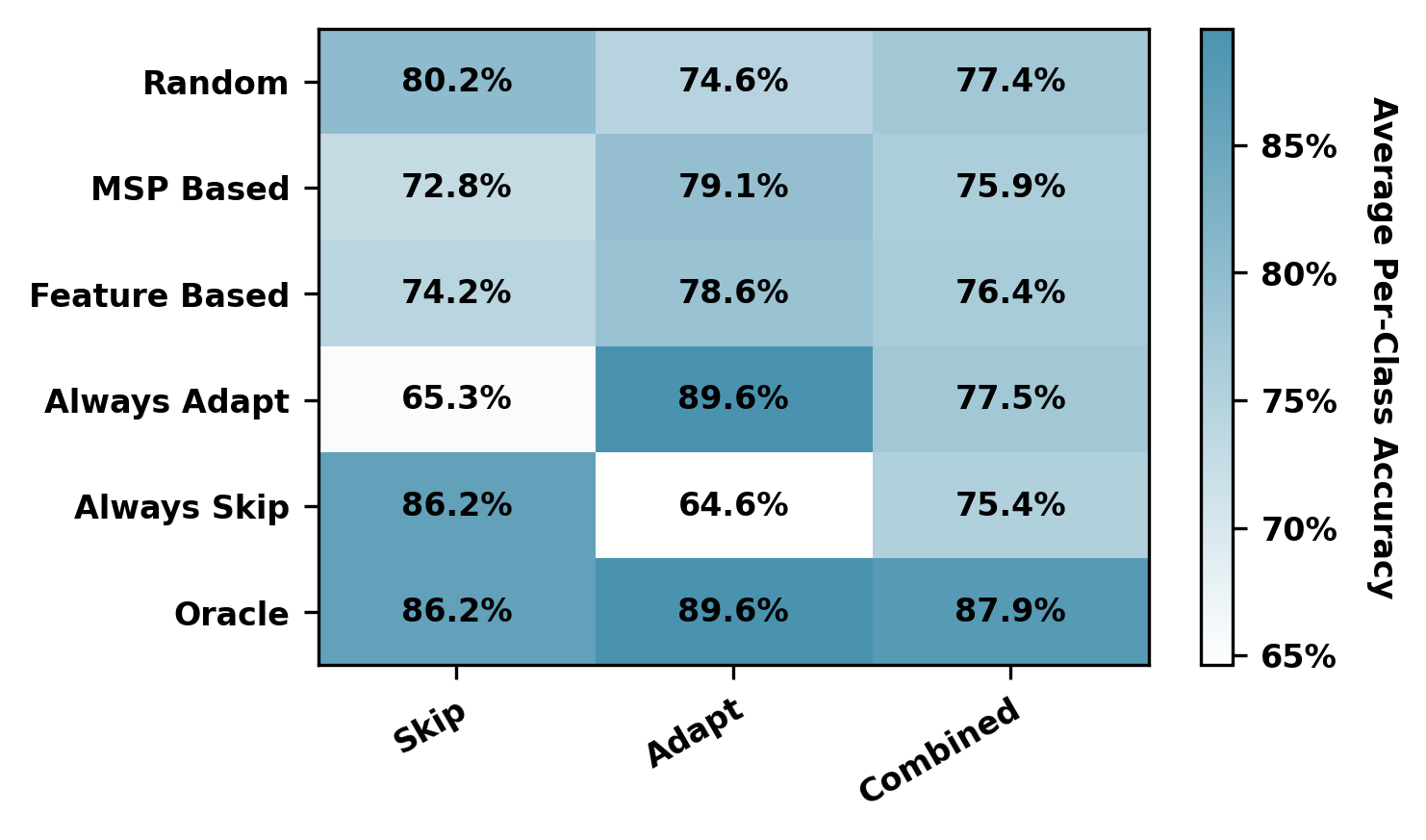}
        \caption{Average Per-Class Accuracy of each method.}
        \label{fig:when2adapt-heatmap-acc}
    \end{subfigure}
    \hfill
    \begin{subfigure}{0.48\linewidth}
        \centering
        \includegraphics[width=\linewidth]{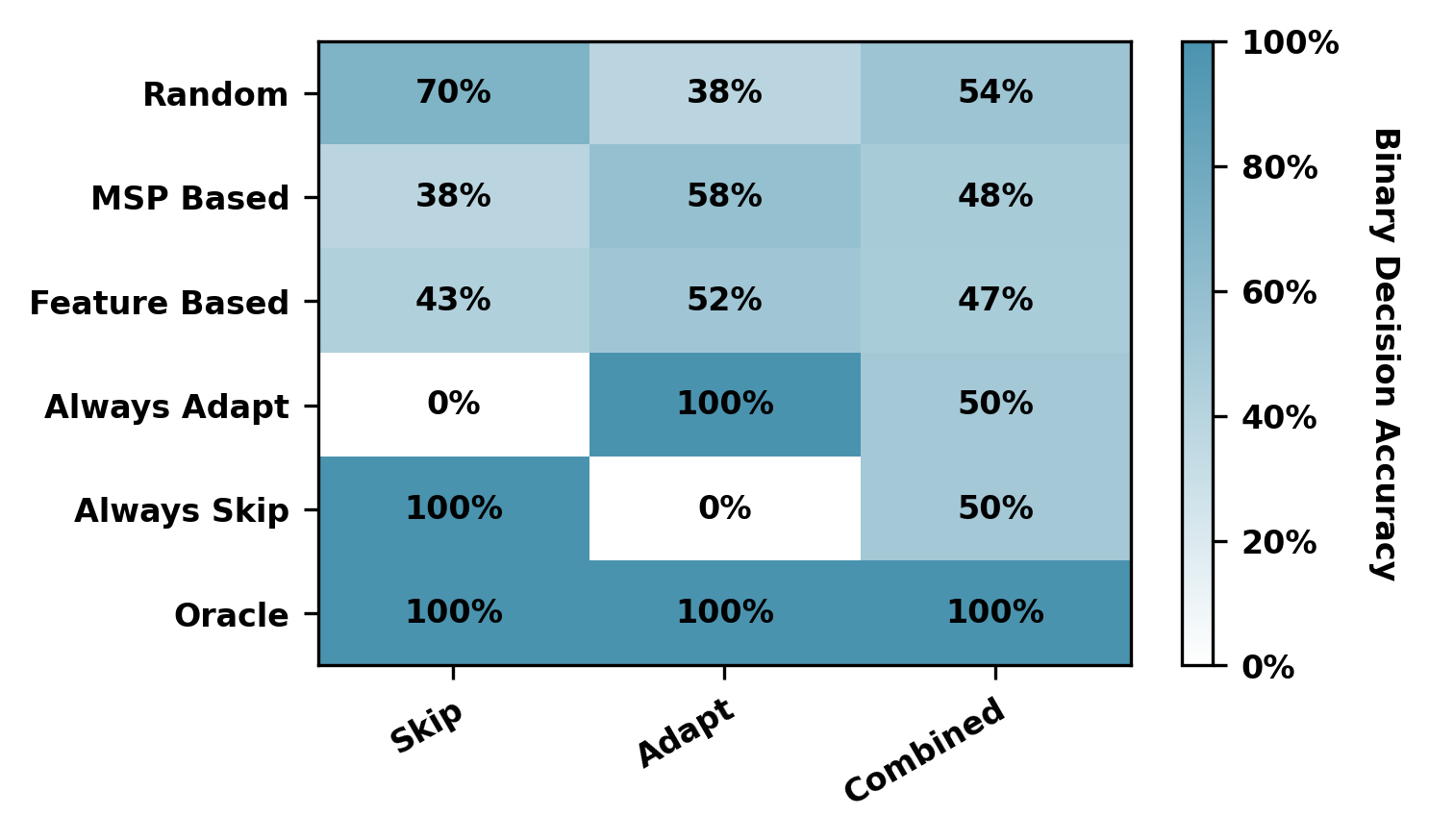}
        \caption{Binary decision accuracy of each method.}
        \label{fig:when2adapt-heatmap-binary}
    \end{subfigure}

    \caption{
    Comparison of adapt-or-not methods on our balanced benchmark. 
    Each interval is labeled as either a Skip case (where not adapting is correct)
    or an Adapt case (where adapting is correct), with Combined reporting 
    overall accuracy across all intervals. Methods include Random, MSP-based, 
    Feature-based, Always Adapt, Always Skip, and an Oracle upper bound. 
    \textbf{(a).} Balanced accuracy under Skip, Adapt, and Combined evaluation. 
    \textbf{(b.)} Binary decision accuracy of each method.
    }
    \label{fig:when2adapt-fig10}
    \vspace{-5mm}
\end{figure}

\noindent\textbf{CLIP feature-distance method.}
We compute class prototypes using labeled data from intervals $1,\dots,j-1$ in CLIP feature space. For images in interval \(j\), we measure the $\ell_2$ distance to the nearest class prototype and average across images. The intuition is that distribution shifts often manifest as changes in the representation space of a model. Adaptation is triggered when the mean distance exceeds threshold $\tau_{\mathrm{feat}}(j)$.

\noindent{Both thresholds are determined using statistics from the zero-shot BioCLIP 2 model applied to the same unlabeled interval.}

\noindent\textbf{Baselines.}
For context, we also include two non-decision baselines: \emph{Always Adapt}, which updates the model at every interval, and \emph{Always Skip}, which never adapts. We additionally report an \emph{Oracle} upper bound that always selects the correct ground-truth action. The ground-truth action is defined as the choice (Adapt or Skip) that yields higher balanced accuracy on the test split of interval $j+1$.

\noindent\textbf{Evaluation metrics.}
We evaluate methods using two metrics: \emph{Balanced Accuracy}, which measures the classification performance obtained by following the method's decisions, and \emph{Binary Decision Accuracy}, which measures the fraction of intervals where the method selects the correct action.

\noindent\textbf{Results and analysis.}
Results are summarized in \cref{fig:when2adapt-heatmap-acc} and \cref{fig:when2adapt-heatmap-binary}. On cases where the ground-truth action is \textbf{Skip}, the oracle and \emph{Always Skip} coincide (\(86.2\%\)), while \emph{Always Adapt} underperforms (\(65.3\%\)). Conversely, in cases where the correct action is \textbf{Adapt}, the oracle and \emph{Always Adapt} coincide (\(89.6\%\), while \emph{Always Skip} falls to \(64.6\%\). The three heuristic decision strategies perform close to random guessing. Their combined balanced accuracies (Random: 77.4\%, MSP: 75.9\%, Feature: 76.4\%) are substantially below the oracle (87.8\%). Binary decision accuracy shows a similar pattern, with MSP and Feature selecting the correct action only 47--48\% of the time.

These results highlight that predicting when adaptation is beneficial remains a challenging and largely unexplored problem, motivating future research on more reliable adapt-or-not decision strategies.

\subsection{More Camera Trap Results}
\label{supp:more_results}
Due to resource constraints, it is not feasible to conduct experiments across all 546 camera traps in \textsc{StreamTrap} for every setting. As such, \cref{tab:all_cameras_new_c_index_rotated} reports the camera traps for which experiments on all settings have been completed. \cref{tab:cam_index_new} summarizes the camera trap index used throughout all figures and tables.

\input{tables_eccv/all_results_new}

\clearpage

\section{\textsc{StreamTrap}: Construction Details}
\label{supp:benchmark}

\subsection{Data Sources}
The LILA BC repository hosts more than fifty datasets and continues to grow. As shown in \cref{tab:dataset_summary}, we limit the benchmark to a subset of camera trap datasets, excluding non-camera-trap sources such as sea-animal imagery, drone imagery, and geological or earth-observation images.

\begin{figure}[tbh]
\small
\centering
\noindent\includegraphics[width=1\linewidth]{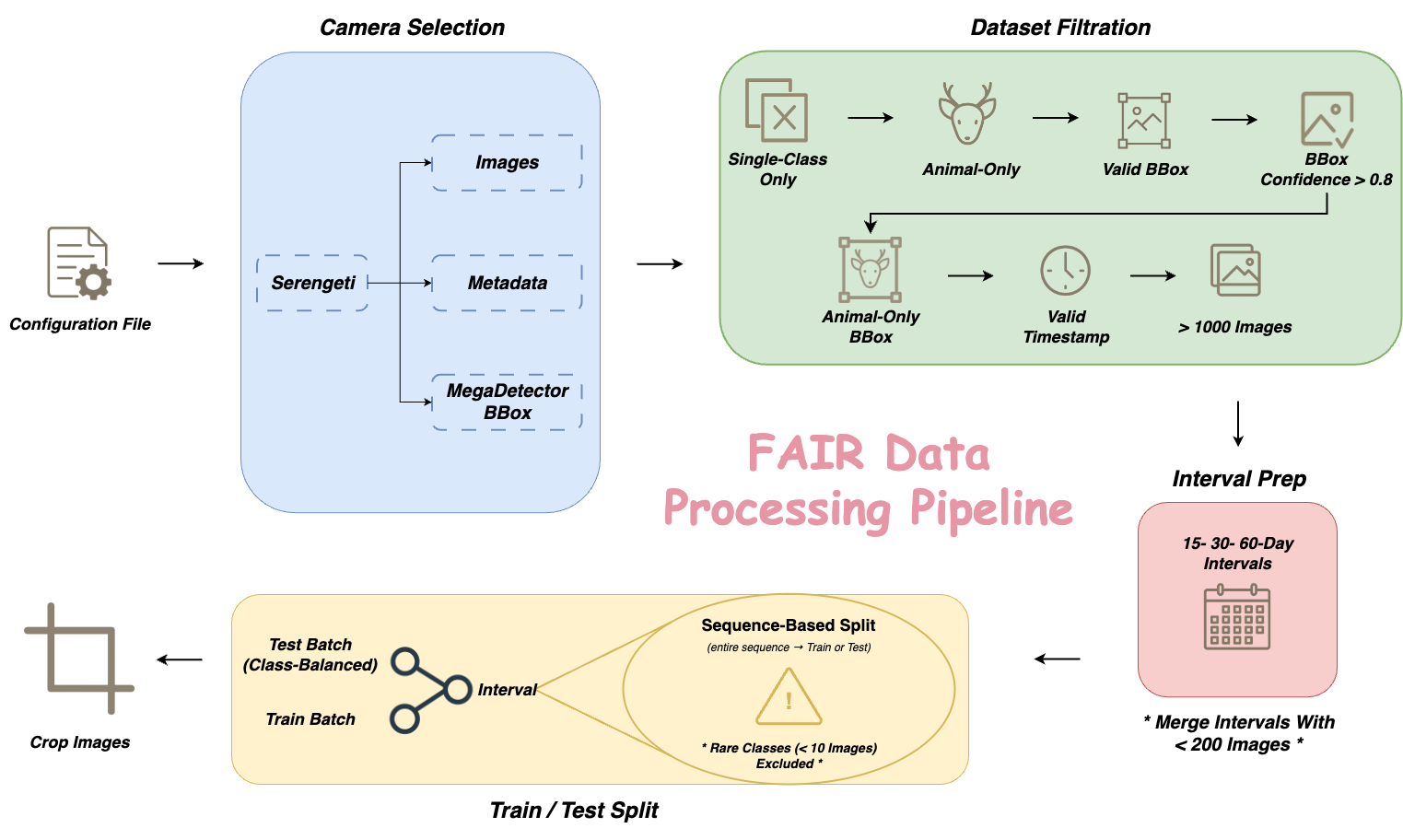}
\vskip -5pt
\caption{\small Detailed camera trap data processing pipeline. Raw camera trap images and metadata are filtered, temporally chunked, and cropped into standardized animal patches, producing per-camera chronological streams for online continual evaluation.}
\label{fig:benchmark_pipeline}
\vspace{-5mm}
\end{figure}

\subsection{Data Processing Pipeline}
The data processing pipeline is driven by a configuration file that specifies the target camera traps, the required filters, and various other experimental settings. Based on this configuration, the pipeline first retrieves and preprocesses the raw data.

\noindent\textbf{Metadata preprocessing.}
All LILA BC camera trap datasets use the COCO camera trap format\footnote{A metadata standard maintained within the LILA BC repository, derived from the original COCO format.} metadata, which includes two optional fields—datetime and sequence—that are essential for our benchmark to group images into temporal intervals and to identify and handle burst images. When datetime is missing, we extract it from each image’s EXIF header\footnote{EXIF stands for Exchangeable Image File Format, a standard for storing metadata such as date and time inside image files.} and insert it into the metadata if available; when sequence is missing, we generate pseudo-sequences by grouping neighboring images within 3 seconds of one another; and when both fields are absent, we first recover datetime from EXIF and then apply the sequence-grouping step. The above metadata preprocessing is implemented as a separate script due to the large amount of image downloading and long processing time it requires; however, it can be incorporated into the pipeline in the future to automatically handle newly added datasets with the same COCO–camera trap format.

\noindent\textbf{MegaDetector bounding box.}
During the initial preprocessing phase, the pipeline also checks the bounding box predictions to ensure they align with the preprocessed metadata. We use MegaDetector~\cite{beery2023megadetector} v5a (MDv5a) as our unified detector across all datasets. MegaDetector evolves through multiple versions, each expanding its training corpus with both private and public camera trap datasets (e.g., Caltech Camera Traps~\cite{beery2018recognition}, Snapshot Serengeti~\cite{swanson2015serengeti}, WCS Camera Traps, NACTI~\cite{tabak2019na}, SWG Camera Traps~\cite{swg2021swg}). MDv5a is trained on all MDv5b training data, along with additional public datasets such as iNaturalist~2017~\cite{van2018inaturalist} and COCO~\cite{lin2014microsoft}. Prior evaluations report strong performance: WildEye reports that MDv5a achieves 99.2\% animal recall at 97.26\% precision, and ecological studies consistently report ~90–98\% recall and 92–99\% precision across diverse environments.

Because several LILA BC datasets now include human-labeled bounding boxes (released after our benchmark preparation) and others still lack such annotations, we apply MDv5a uniformly to all 17 datasets to ensure consistent preprocessing. For datasets without human-labeled boxes, we rely entirely on MDv5a detections; for those with human labels, we still apply MDv5a with a high-confidence threshold for consistency. MDv5a outputs bounding boxes for animals, humans, and vehicles; we retain only animal detections and use metadata to link each bounding box to its species. Our goal is not to study detection itself but to build a consistent and reliable foundation for classification.

\noindent\textbf{Data processing.}
\label{ss:Data Processing}
Once the initial data retrieval and metadata-detector matching are complete, the finalized data is passed through our processing pipeline, illustrated in \cref{fig:benchmark_pipeline}. Applying the settings from the configuration file, the pipeline filters the data based on dataset or camera selection, minimum image count constraints, bounding-box confidence thresholds, and class exclusion for targeted taxa. Low-quality and other hard cases are filtered through these settings, enabling control over benchmark difficulty and data quality. This filtering step produces a substantial reduction in both images and camera traps, as summarized in \cref{tab:dataset_summary}, with 80.1\% of images and 88.9\% of camera traps filtered overall. This is largely driven by the prevalence of empty images: approximately 41.7\% contain no animals. Such cases naturally arise in camera trap systems, where motion triggers frequently activate after animals leave the scene or due to minor environmental disturbances.

After removing non-informative images, the remaining valid data is grouped into temporal intervals, with a default window of 30-day chunks (adjustable to 15, 60, or custom durations). To ensure sufficient data for training and testing, we merge intervals when a chunk contains fewer than 200 images. Within each interval, the data is partitioned into training and testing sets. Crucially, this split is performed while ensuring that burst images within the same sequence are kept in the same set to prevent temporal data leakage. While a fixed seed is not enforced by default, it can be specified to ensure reproducible splits, and rare class interval data may optionally be preserved alongside the training and test sets to support long-tail evaluation. Finally, the pipeline crops the actual images. Following prior work, we enlarge each detected region by 50\% of the bounding box size before cropping. The pipeline then saves the prepared interval JSON data alongside these cropped patches as the final benchmark output for each camera trap. The processed camera trap list is summarized in \cref{tab:camera_summary}.

\begin{figure*}[t]
\small
\centering
\noindent\includegraphics[width=1\linewidth]{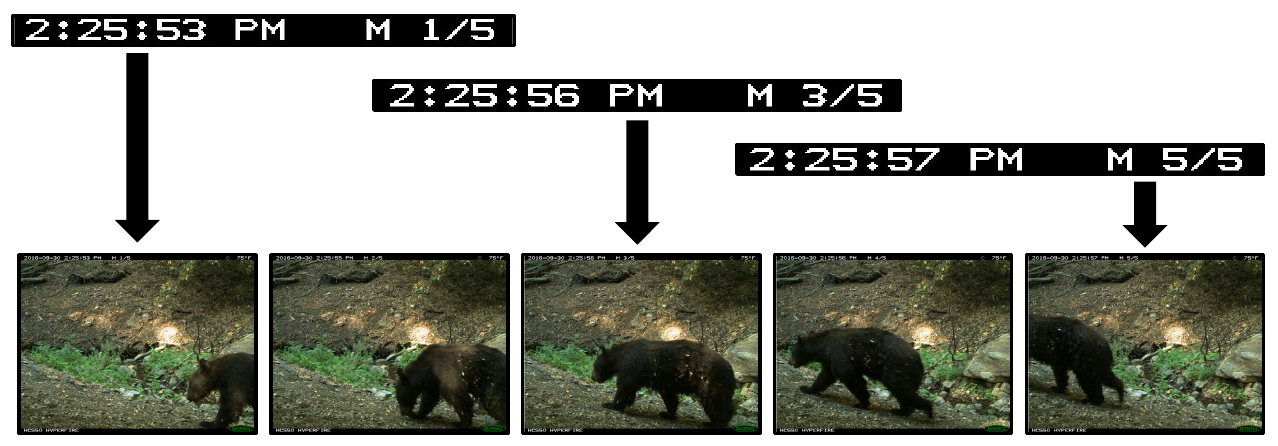}
\vskip -5pt
\caption{\small Illustration of burst images from a single sequence. Upon detecting motion, the camera captures multiple frames in rapid succession, producing images that share the same environment, background, and animal appearance.}
\vskip -10pt
\label{fig:burst}
\end{figure*}

\subsection{One Trigger, Multiple Frames: Burst Images}
\label{sec:add_details_data_process}

Camera trap devices capture bursts of images when motion is detected, with each burst containing 3 to over 10 frames, which we refer to as a \emph{sequence}. Using sequence IDs and frame-order metadata, we assign all frames from each sequence to a single split (train or test). As shown in \cref{fig:burst}, frames within a sequence are captured within seconds, share nearly identical backgrounds and viewpoints, and depict the same animal, resulting in extremely high visual correlation. Splitting such frames across train and test would therefore introduce severe temporal leakage, artificially inflating performance by allowing near-duplicate content to appear in both splits. To avoid this, we enforce strict sequence-level splitting.

While splits are defined at the sequence level, each image is treated as an independent training and evaluation sample. We acknowledge that explicitly modeling intra-sequence correlations may further improve performance, but it is beyond our current scope. Moreover, evaluating at the sequence level, via single-frame downsampling or aggregating predictions across sequences, could provide a more realistic and potentially more challenging setting. Quantifying the resulting dataset size reduction and performance shifts remains an important direction for future work.




\input{tables_eccv/dataset_table}
\input{tables_eccv/camera_trap_index_new}
\input{tables_eccv/vidhi_cam_list}

%% file: tables_eccv/zs_comparison.tex
\begin{table}[h!]
    \vspace{3mm}
    \centering
    \begin{tabular}{cc}
        \toprule
        BioCLIP 2~\cite{gu2025bioclip} & CLIP~\cite{openaiCLIP} \\
        \midrule
        84.3 & 72.3 \\
        \bottomrule
    \end{tabular}
    \vspace{3mm}
    \caption{\small{Zero-shot baseline selection. Average per-class accuracy (\%) across all camera traps in the \textsc{StreamTrap}. BioCLIP~2 is selected as the zero-shot baseline due to its superior performance.}}
    \vspace{-8mm}
    \label{tab:zs_comparison}
\end{table}

%% file: tables_eccv/train_details.tex
\begin{table}[tbh]
\vspace{3mm}
\centering
\small
\begin{tabular}{cc}
\toprule
\textbf{Hyperparameter} & \textbf{Value} \\
\midrule
Optimizer & AdamW \\
Max Learning Rate & $2.5\times10^{-5}$ \\
Min Learning Rate & $4.17\times10^{-7}$ \\
Weight Decay & $1\times10^{-4}$ \\
LR Schedule & Cosine Annealing ($T_{\text{max}}=60$) \\
Batch Size & 32 \\
\bottomrule
\end{tabular}
\vspace{3mm}
\caption{Training configuration for fine-tuning BioCLIP~2.}
\label{tab:training_config}
\vspace{-8mm}
\end{table}

%% file: tables_eccv/oracle_loss_peft.tex
\begin{table}[tbh]
\vspace{-8mm}
\centering
\setlength{\tabcolsep}{10pt}
\begin{tabular}{l l c}
\toprule
Loss & FT Method & Accuracy \\
\midrule
\rowcolor{gray!15}
CE        & Full & 85.9 \\
\midrule
BSM~\cite{ren2020balanced}       & Full & \textbf{85.1} \\
CB-Focal~\cite{cui2019class}  & Full & 83.2 \\
CDT~\cite{ye2020identifying}       & Full & 83.2 \\
\midrule
CE        & LoRA~\cite{hu2022lora}    & \textbf{86.0} \\
CE        & VPT~\cite{vpt}     & 76.6 \\
CE        & Adapter~\cite{houlsbyadapter} & 85.3 \\
\midrule
\rowcolor{blue!10} BSM~\cite{ren2020balanced}        & LoRA~\cite{hu2022lora}     & \textbf{90.3} \\
\bottomrule
\end{tabular}
\vspace{3mm}
\caption{\small{Ablation of imbalance losses and PEFT fine-tuning (FT) methods. Average per-class accuracy (\%) is reported. BSM and LoRA achieve the best performance. Our proposed adaptation recipe ($\star$) is highlighted in \textcolor{recipeblue}{\textbf{blue}}.}}
\vspace{-8mm}
\label{tab:loss_peft}
\end{table}

%% file: tables_eccv/openset_and_rare.tex
\begin{table}[tbh]
    \centering
    \footnotesize
    \setlength{\tabcolsep}{5pt}
    \begin{tabular}{lccc}
        \toprule
         & Closed-Set & Open-Set & Closed-Set + Rare Species \\
        \midrule
        Zero-shot & 82.5 & 37.8 & 73.1 \\
        Accum$\star$ & 83.7 & 76.7 & 73.4 \\
        Oracle$\star$ & 88.3 & 86.1 & 66.8 \\
        \bottomrule
    \end{tabular}
    \vspace{3mm}
    \caption{\small Average per-class accuracy (\%) under relaxed evaluation assumptions. Open-Set expands the vocabulary to all 1,096 classes across \textsc{StreamTrap}. Rare Species adds rare classes (fewer than 10 samples per interval) to the test set.}
    \label{tab:openset_and_rare}
    \vspace{-8mm}
\end{table}

%% file: tables_eccv/appendix_post_process.tex
\begin{table}[tbh]
    \centering
    \footnotesize
    \setlength{\tabcolsep}{5pt}
    \begin{tabular}{lcc}
        \toprule
         & Accuracy & $\Delta{\text{Oracle}\star}$ \\
        \midrule
        Oracle$\star$ & 87.0 & -- \\
Accum$\star$ & 79.8 & 7.1 \\
\quad + Calibration & 83.7 & 3.3 \\
\quad + Weight Interpolation & 83.2 & 3.8 \\
\quad + Model Selection & 86.1 & 0.9 \\
\rowcolor{blue!10} \quad + Best Post-Processing & 86.7 & 0.2 \\
        \bottomrule
    \end{tabular}
    \vspace{3mm}
    \caption{Extended post-processing results by average per-class accuracy (\%). $\Delta{\text{oracle}\star}$ denotes remaining gap to Oracle$\star$ (87.6 \%). Each method is applied on top of accum$\star$. The \textcolor{blue!45}{\textbf{last}} row selects the best-performing post-processing method per camera trap.}
    \label{tab:postproc}
    \vspace{-12mm}
\end{table}

%% file: tables_eccv/all_results_new.tex
\begin{sidewaystable*}
\centering
\scriptsize
\setlength{\tabcolsep}{2pt}
\renewcommand{\arraystretch}{1.5}
\setlength{\aboverulesep}{0pt}
\setlength{\belowrulesep}{0pt}

\resizebox{\linewidth}{!}{
\begin{tabular}{@{\hspace{2pt}}@{\extracolsep{\fill}} l g w g w g w g w g w g w g w g w g w g w g w g w g w g w g w g w g w g w g w g w @{}}
\toprule
 & {C1} & {C2} & {C3} & {C4} & {C5} & {C6} & {C7} & {C8} & {C9} & {C10} & {C11} & {C12} & {C13} & {C14} & {C15} & {C16} & {C17} & {C18} & {C19} & {C20} & {C21} & {C22} & {C23} & {C24} & {C25} & {C26} & {C27} & {C28} & {C29} & {C30} & {C31} & {C32} & {C33} & {C34} & {C35} & {C36} & {C37} & {C38} & {C39} & Avg \\
\midrule
ZS & 97.7 & 92.3 & 90.9 & 89.1 & 86.1 & 84.3 & 81.5 & 81.2 & 78.3 & 77.1 & 75.8 & 73.8 & 70.1 & 69.8 & 67.1 & 66.0 & 62.2 & 61.5 & 56.3 & 55.8 & 94.5 & 94.1 & 90.9 & 89.0 & 87.2 & 85.7 & 85.6 & 85.3 & 83.1 & 83.0 & 82.6 & 82.1 & 81.6 & 81.5 & 81.3 & 81.0 & 80.9 & 80.4 & 80.2 & 80.2 \\
\midrule
Accum & 73.6 & 77.1 & 50.0 & 71.7 & 46.7 & 33.3 & 56.7 & 73.0 & 45.0 & 25.0 & 57.0 & 50.0 & 68.2 & 60.8 & 66.7 & 11.4 & 61.7 & 58.0 & 33.3 & 45.7 & 72.9 & 63.2 & 53.6 & 62.3 & 75.1 & 90.0 & 75.7 & 26.3 & 62.5 & 54.2 & 68.0 & 72.0 & 96.7 & 50.0 & 56.0 & 36.8 & 72.3 & 72.3 & 74.0 & 58.9 \\
$\Delta$ & \color{negred} 24.1 & \color{negred} 15.2 & \color{negred} 40.9 & \color{negred} 17.4 & \color{negred} 39.4 & \color{negred} 51.0 & \color{negred} 24.8 & \color{negred} 8.2 & \color{negred} 33.3 & \color{negred} 52.1 & \color{negred} 18.8 & \color{negred} 23.8 & \color{negred} 1.9 & \color{negred} 9.0 & \color{negred} 0.4 & \color{negred} 54.6 & \color{negred} 0.5 & \color{negred} 3.5 & \color{negred} 23.0 & \color{negred} 10.1 & \color{negred} 21.6 & \color{negred} 30.9 & \color{negred} 37.3 & \color{negred} 26.7 & \color{negred} 12.1 & \color{posblue} 4.3 & \color{negred} 9.9 & \color{negred} 59.0 & \color{negred} 20.6 & \color{negred} 28.8 & \color{negred} 14.6 & \color{negred} 10.1 & \color{posblue} 15.1 & \color{negred} 31.5 & \color{negred} 25.3 & \color{negred} 44.2 & \color{negred} 8.6 & \color{negred} 8.1 & \color{negred} 6.2 & \multicolumn{1}{c}{--} \\
\midrule
Accum$\star$ & 98.7 & 89.9 & 89.9 & 90.8 & 79.4 & 79.2 & 83.4 & 88.3 & 81.0 & 78.7 & 76.9 & 70.1 & 79.4 & 75.4 & 86.9 & 76.6 & 88.9 & 71.3 & 76.7 & 82.6 & 94.7 & 98.0 & 92.2 & 86.6 & 88.2 & 82.1 & 94.5 & 79.1 & 79.5 & 86.4 & 81.0 & 83.0 & 81.5 & 69.7 & 79.6 & 71.8 & 87.2 & 83.5 & 85.4 & 83.3 \\
$\Delta$ & \color{posblue} 1.0 & \color{negred} 2.4 & \color{negred} 1.0 & \color{posblue} 1.7 & \color{negred} 6.7 & \color{negred} 5.1 & \color{posblue} 1.9 & \color{posblue} 7.1 & \color{posblue} 2.7 & \color{posblue} 1.6 & \color{posblue} 1.1 & \color{negred} 3.7 & \color{posblue} 9.3 & \color{posblue} 5.6 & \color{posblue} 19.8 & \color{posblue} 10.6 & \color{posblue} 26.7 & \color{posblue} 9.8 & \color{posblue} 20.4 & \color{posblue} 26.8 & \color{posblue} 0.2 & \color{posblue} 3.9 & \color{posblue} 1.3 & \color{negred} 2.4 & \color{posblue} 1.0 & \color{negred} 3.6 & \color{posblue} 8.9 & \color{negred} 6.2 & \color{negred} 3.6 & \color{posblue} 3.4 & \color{negred} 1.6 & \color{posblue} 0.9 & \color{negred} 0.1 & \color{negred} 11.8 & \color{negred} 1.7 & \color{negred} 9.2 & \color{posblue} 6.3 & \color{posblue} 3.1 & \color{posblue} 5.2 & \multicolumn{1}{c}{--} \\
\midrule
{\color{gray} Oracle$\star$} & \color{gray} 98.7 & \color{gray} 96.4 & \color{gray} 94.5 & \color{gray} 96.6 & \color{gray} 87.8 & \color{gray} 91.5 & \color{gray} 91.0 & \color{gray} 91.0 & \color{gray} 87.3 & \color{gray} 82.0 & \color{gray} 84.3 & \color{gray} 88.5 & \color{gray} 91.0 & \color{gray} 86.6 & \color{gray} 91.0 & \color{gray} 83.9 & \color{gray} 93.8 & \color{gray} 89.6 & \color{gray} 89.9 & \color{gray} 93.4 & \color{gray} 96.5 & \color{gray} 96.1 & \color{gray} 95.8 & \color{gray} 91.9 & \color{gray} 90.3 & \color{gray} 88.6 & \color{gray} 90.0 & \color{gray} 87.7 & \color{gray} 89.9 & \color{gray} 96.3 & \color{gray} 87.8 & \color{gray} 86.8 & \color{gray} 86.9 & \color{gray} 82.2 & \color{gray} 84.9 & \color{gray} 83.4 & \color{gray} 91.9 & \color{gray} 91.9 & \color{gray} 87.8 & \color{gray} 90.1 \\
$\Delta$ & \color{posblue} 1.0 & \color{posblue} 4.1 & \color{posblue} 3.6 & \color{posblue} 7.5 & \color{posblue} 1.7 & \color{posblue} 7.2 & \color{posblue} 9.5 & \color{posblue} 9.8 & \color{posblue} 9.0 & \color{posblue} 4.9 & \color{posblue} 8.5 & \color{posblue} 14.7 & \color{posblue} 20.9 & \color{posblue} 16.8 & \color{posblue} 23.9 & \color{posblue} 17.9 & \color{posblue} 31.6 & \color{posblue} 28.1 & \color{posblue} 33.6 & \color{posblue} 37.6 & \color{posblue} 2.0 & \color{posblue} 2.0 & \color{posblue} 4.9 & \color{posblue} 2.9 & \color{posblue} 3.1 & \color{posblue} 2.9 & \color{posblue} 4.4 & \color{posblue} 2.4 & \color{posblue} 6.8 & \color{posblue} 13.3 & \color{posblue} 5.2 & \color{posblue} 4.7 & \color{posblue} 5.3 & \color{posblue} 0.7 & \color{posblue} 3.6 & \color{posblue} 2.4 & \color{posblue} 11.0 & \color{posblue} 11.5 & \color{posblue} 7.6 & \multicolumn{1}{c}{--} \\
\bottomrule
\end{tabular}
}

\vspace{3mm}

\resizebox{\linewidth}{!}{
\begin{tabular}{@{\hspace{2pt}}@{\extracolsep{\fill}} l g w g w g w g w g w g w g w g w g w g w g w g w g w g w g w g w g w g w g w g w @{}}
\toprule
 & {C40} & {C41} & {C42} & {C43} & {C44} & {C45} & {C46} & {C47} & {C48} & {C49} & {C50} & {C51} & {C52} & {C53} & {C54} & {C55} & {C56} & {C57} & {C58} & {C59} & {C60} & {C61} & {C62} & {C63} & {C64} & {C65} & {C66} & {C67} & {C68} & {C69} & {C70} & {C71} & {C72} & {C73} & {C74} & {C75} & {C76} & {C77} & {C78} & Avg \\
\midrule
ZS & 80.1 & 79.7 & 79.3 & 79.1 & 79.0 & 78.9 & 78.3 & 77.9 & 77.5 & 76.6 & 76.5 & 75.8 & 75.5 & 75.3 & 74.7 & 72.2 & 72.0 & 71.4 & 69.6 & 69.5 & 67.6 & 66.7 & 66.0 & 64.7 & 64.1 & 63.2 & 62.7 & 59.8 & 58.6 & 58.3 & 56.9 & 98.0 & 97.3 & 96.3 & 95.9 & 95.8 & 95.8 & 94.1 & 94.0 & 76.3 \\
\midrule
Accum & 60.0 & 27.1 & 76.0 & 25.0 & 33.3 & 50.0 & 56.2 & 73.0 & 50.0 & 70.9 & 68.0 & 42.5 & 38.0 & 66.0 & 54.0 & 72.3 & 20.0 & 63.9 & 16.7 & 66.0 & 25.0 & 54.3 & 45.0 & 56.9 & 26.0 & 39.7 & 60.6 & 64.3 & 66.6 & 16.7 & 50.0 & 81.3 & 89.4 & 53.6 & 76.8 & 71.4 & 81.7 & 90.4 & 64.2 & 54.9 \\
$\Delta$ & \color{negred} 20.1 & \color{negred} 52.6 & \color{negred} 3.3 & \color{negred} 54.1 & \color{negred} 45.7 & \color{negred} 28.9 & \color{negred} 22.1 & \color{negred} 4.9 & \color{negred} 27.5 & \color{negred} 5.7 & \color{negred} 8.5 & \color{negred} 33.3 & \color{negred} 37.5 & \color{negred} 9.3 & \color{negred} 20.7 & \color{posblue} 0.1 & \color{negred} 52.0 & \color{negred} 7.5 & \color{negred} 52.9 & \color{negred} 3.5 & \color{negred} 42.6 & \color{negred} 12.4 & \color{negred} 21.0 & \color{negred} 7.8 & \color{negred} 38.1 & \color{negred} 23.5 & \color{negred} 2.1 & \color{posblue} 4.5 & \color{posblue} 8.0 & \color{negred} 41.6 & \color{negred} 6.9 & \color{negred} 16.7 & \color{negred} 7.9 & \color{negred} 42.7 & \color{negred} 19.1 & \color{negred} 24.4 & \color{negred} 14.1 & \color{negred} 3.7 & \color{negred} 29.8 & \multicolumn{1}{c}{--} \\
\midrule
Accum$\star$ & 75.9 & 70.3 & 90.7 & 78.1 & 82.2 & 78.3 & 77.7 & 84.9 & 86.2 & 77.9 & 80.9 & 79.4 & 77.2 & 80.0 & 71.8 & 83.3 & 72.7 & 72.6 & 66.4 & 79.6 & 75.0 & 81.6 & 66.5 & 65.3 & 81.0 & 83.4 & 75.4 & 75.7 & 68.7 & 79.3 & 70.0 & 98.9 & 97.7 & 95.8 & 95.9 & 96.7 & 95.1 & 90.2 & 92.0 & 80.8 \\
$\Delta$ & \color{negred} 4.2 & \color{negred} 9.4 & \color{posblue} 11.4 & \color{negred} 1.0 & \color{posblue} 3.2 & \color{negred} 0.6 & \color{negred} 0.6 & \color{posblue} 7.0 & \color{posblue} 8.7 & \color{posblue} 1.3 & \color{posblue} 4.4 & \color{posblue} 3.6 & \color{posblue} 1.7 & \color{posblue} 4.7 & \color{negred} 2.9 & \color{posblue} 11.1 & \color{posblue} 0.7 & \color{posblue} 1.2 & \color{negred} 3.2 & \color{posblue} 10.1 & \color{posblue} 7.4 & \color{posblue} 14.9 & \color{posblue} 0.5 & \color{posblue} 0.6 & \color{posblue} 16.9 & \color{posblue} 20.2 & \color{posblue} 12.7 & \color{posblue} 15.9 & \color{posblue} 10.1 & \color{posblue} 21.0 & \color{posblue} 13.1 & \color{posblue} 0.9 & \color{posblue} 0.4 & \color{negred} 0.5 & \color{posblue} 0.0 & \color{posblue} 0.9 & \color{negred} 0.7 & \color{negred} 3.9 & \color{negred} 2.0 & \multicolumn{1}{c}{--} \\
\midrule
{\color{gray} Oracle$\star$} & \color{gray} 81.6 & \color{gray} 87.9 & \color{gray} 92.1 & \color{gray} 84.0 & \color{gray} 86.9 & \color{gray} 84.8 & \color{gray} 84.1 & \color{gray} 91.6 & \color{gray} 89.4 & \color{gray} 87.4 & \color{gray} 87.0 & \color{gray} 83.3 & \color{gray} 81.3 & \color{gray} 82.4 & \color{gray} 84.3 & \color{gray} 93.3 & \color{gray} 79.8 & \color{gray} 86.4 & \color{gray} 87.0 & \color{gray} 78.7 & \color{gray} 85.0 & \color{gray} 91.6 & \color{gray} 91.3 & \color{gray} 70.3 & \color{gray} 89.9 & \color{gray} 84.7 & \color{gray} 89.7 & \color{gray} 80.4 & \color{gray} 79.4 & \color{gray} 81.4 & \color{gray} 85.4 & \color{gray} 99.8 & \color{gray} 98.2 & \color{gray} 97.8 & \color{gray} 98.3 & \color{gray} 98.3 & \color{gray} 96.3 & \color{gray} 94.9 & \color{gray} 94.3 & \color{gray} 87.7 \\
$\Delta$ & \color{posblue} 1.5 & \color{posblue} 8.2 & \color{posblue} 12.8 & \color{posblue} 4.9 & \color{posblue} 7.9 & \color{posblue} 5.9 & \color{posblue} 5.8 & \color{posblue} 13.7 & \color{posblue} 11.9 & \color{posblue} 10.8 & \color{posblue} 10.5 & \color{posblue} 7.5 & \color{posblue} 5.8 & \color{posblue} 7.1 & \color{posblue} 9.6 & \color{posblue} 21.1 & \color{posblue} 7.8 & \color{posblue} 15.0 & \color{posblue} 17.4 & \color{posblue} 9.2 & \color{posblue} 17.4 & \color{posblue} 24.9 & \color{posblue} 25.3 & \color{posblue} 5.6 & \color{posblue} 25.8 & \color{posblue} 21.5 & \color{posblue} 27.0 & \color{posblue} 20.6 & \color{posblue} 20.8 & \color{posblue} 23.1 & \color{posblue} 28.5 & \color{posblue} 1.8 & \color{posblue} 0.9 & \color{posblue} 1.5 & \color{posblue} 2.4 & \color{posblue} 2.5 & \color{posblue} 0.5 & \color{posblue} 0.8 & \color{posblue} 0.3 & \multicolumn{1}{c}{--} \\
\bottomrule
\end{tabular}
}

\vspace{3mm}

\resizebox{\linewidth}{!}{
\begin{tabular}{@{\hspace{2pt}}@{\extracolsep{\fill}} l g w g w g w g w g w g w g w g w g w g w g w g w g w g w g w g w g w g w g w w @{}}
\toprule
 & {C79} & {C80} & {C81} & {C82} & {C83} & {C84} & {C85} & {C86} & {C87} & {C88} & {C89} & {C90} & {C91} & {C92} & {C93} & {C94} & {C95} & {C96} & {C97} & {C98} & {C99} & {C100} & {C101} & {C102} & {C103} & {C104} & {C105} & {C106} & {C107} & {C108} & {C109} & {C110} & {C111} & {C112} & {C113} & {C114} & {C115} & {C116} & Avg \\
\midrule
ZS & 93.9 & 93.4 & 93.4 & 93.3 & 92.8 & 91.9 & 91.3 & 90.7 & 90.2 & 89.9 & 89.6 & 88.3 & 88.0 & 87.7 & 87.7 & 86.8 & 84.7 & 84.6 & 82.4 & 82.0 & 79.8 & 79.3 & 78.9 & 78.4 & 75.6 & 75.6 & 74.8 & 74.8 & 73.9 & 71.9 & 71.2 & 70.8 & 70.4 & 67.0 & 66.7 & 63.6 & 61.8 & 60.6 & 81.0 \\
\midrule
Accum & 82.0 & 81.0 & 78.0 & 62.3 & 88.0 & 83.0 & 70.8 & 78.2 & 78.1 & 82.7 & 84.7 & 83.2 & 77.0 & 88.3 & 72.2 & 69.1 & 83.5 & 81.1 & 78.5 & 75.5 & 75.8 & 79.3 & 84.5 & 71.5 & 79.0 & 73.3 & 67.5 & 69.5 & 65.4 & 66.2 & 61.8 & 75.8 & 65.7 & 61.4 & 64.8 & 56.0 & 64.2 & 79.2 & 74.7 \\
$\Delta$ & \color{negred} 11.9 & \color{negred} 12.4 & \color{negred} 15.4 & \color{negred} 31.0 & \color{negred} 4.8 & \color{negred} 8.9 & \color{negred} 20.5 & \color{negred} 12.5 & \color{negred} 12.1 & \color{negred} 7.2 & \color{negred} 4.9 & \color{negred} 5.1 & \color{negred} 11.0 & \color{posblue} 0.6 & \color{negred} 15.5 & \color{negred} 17.7 & \color{negred} 1.2 & \color{negred} 3.5 & \color{negred} 3.9 & \color{negred} 6.5 & \color{negred} 4.0 & \color{posblue} 0.0 & \color{posblue} 5.6 & \color{negred} 6.9 & \color{posblue} 3.4 & \color{negred} 2.3 & \color{negred} 7.3 & \color{negred} 5.3 & \color{negred} 8.5 & \color{negred} 5.7 & \color{negred} 9.4 & \color{posblue} 5.0 & \color{negred} 4.7 & \color{negred} 5.6 & \color{negred} 1.9 & \color{negred} 7.6 & \color{posblue} 2.4 & \color{posblue} 18.6 & \multicolumn{1}{c}{--} \\
\midrule
Accum$\star$ & 93.1 & 93.8 & 94.2 & 92.9 & 92.8 & 88.1 & 92.5 & 88.1 & 94.4 & 92.5 & 93.2 & 90.9 & 90.2 & 90.8 & 87.0 & 84.8 & 85.5 & 86.0 & 90.3 & 88.1 & 88.4 & 89.3 & 91.1 & 88.5 & 70.0 & 80.4 & 76.7 & 74.2 & 61.6 & 78.2 & 81.5 & 79.8 & 86.8 & 61.5 & 77.3 & 33.4 & 84.2 & 87.6 & 84.2 \\
$\Delta$ & \color{negred} 0.8 & \color{posblue} 0.4 & \color{posblue} 0.8 & \color{negred} 0.4 & \color{posblue} 0.0 & \color{negred} 3.8 & \color{posblue} 1.2 & \color{negred} 2.6 & \color{posblue} 4.2 & \color{posblue} 2.6 & \color{posblue} 3.6 & \color{posblue} 2.6 & \color{posblue} 2.2 & \color{posblue} 3.1 & \color{negred} 0.7 & \color{negred} 2.0 & \color{posblue} 0.8 & \color{posblue} 1.4 & \color{posblue} 7.9 & \color{posblue} 6.1 & \color{posblue} 8.6 & \color{posblue} 10.0 & \color{posblue} 12.2 & \color{posblue} 10.1 & \color{negred} 5.6 & \color{posblue} 4.8 & \color{posblue} 1.9 & \color{negred} 0.6 & \color{negred} 12.3 & \color{posblue} 6.3 & \color{posblue} 10.3 & \color{posblue} 9.0 & \color{posblue} 16.4 & \color{negred} 5.5 & \color{posblue} 10.6 & \color{negred} 30.2 & \color{posblue} 22.4 & \color{posblue} 27.0 & \multicolumn{1}{c}{--} \\
\midrule
{\color{gray} Oracle$\star$} & \color{gray} 95.7 & \color{gray} 96.5 & \color{gray} 96.9 & \color{gray} 94.9 & \color{gray} 98.4 & \color{gray} 95.5 & \color{gray} 95.0 & \color{gray} 92.4 & \color{gray} 95.6 & \color{gray} 95.9 & \color{gray} 94.8 & \color{gray} 95.2 & \color{gray} 92.4 & \color{gray} 94.0 & \color{gray} 92.0 & \color{gray} 91.9 & \color{gray} 94.0 & \color{gray} 87.7 & \color{gray} 93.4 & \color{gray} 91.3 & \color{gray} 91.2 & \color{gray} 91.9 & \color{gray} 82.6 & \color{gray} 90.7 & \color{gray} 88.5 & \color{gray} 83.0 & \color{gray} 88.2 & \color{gray} 79.4 & \color{gray} 83.2 & \color{gray} 80.4 & \color{gray} 80.1 & \color{gray} 84.8 & \color{gray} 90.2 & \color{gray} 83.8 & \color{gray} 79.6 & \color{gray} 76.3 & \color{gray} 86.9 & \color{gray} 96.7 & \color{gray} 90.0 \\
$\Delta$ & \color{posblue} 1.8 & \color{posblue} 3.1 & \color{posblue} 3.5 & \color{posblue} 1.6 & \color{posblue} 5.6 & \color{posblue} 3.6 & \color{posblue} 3.7 & \color{posblue} 1.7 & \color{posblue} 5.4 & \color{posblue} 6.0 & \color{posblue} 5.2 & \color{posblue} 6.9 & \color{posblue} 4.4 & \color{posblue} 6.3 & \color{posblue} 4.3 & \color{posblue} 5.1 & \color{posblue} 9.3 & \color{posblue} 3.1 & \color{posblue} 11.0 & \color{posblue} 9.3 & \color{posblue} 11.4 & \color{posblue} 12.6 & \color{posblue} 3.7 & \color{posblue} 12.3 & \color{posblue} 12.9 & \color{posblue} 7.4 & \color{posblue} 13.4 & \color{posblue} 4.6 & \color{posblue} 9.3 & \color{posblue} 8.5 & \color{posblue} 8.9 & \color{posblue} 14.0 & \color{posblue} 19.8 & \color{posblue} 16.8 & \color{posblue} 12.9 & \color{posblue} 12.7 & \color{posblue} 25.1 & \color{posblue} 36.1 & \multicolumn{1}{c}{--} \\
\bottomrule
\end{tabular}
}

\vspace{2mm}
\caption{\small {Complete result of ZS, Oracle$\star$, Accum, Accum$\star$.} \textcolor{posblue}{\textbf{Blue}} (\textcolor{negred}{\textbf{red}}) denotes improvement (degradation) relative to ZS.}
\label{tab:all_cameras_new_c_index_rotated}
\end{sidewaystable*}

%% file: tables_eccv/dataset_table.tex
\begin{table*}[tbh]
\centering
\scriptsize
\setlength{\tabcolsep}{3.5pt}
\renewcommand{\arraystretch}{1.1}
\resizebox{\textwidth}{!}{%
\begin{tabular}{l l r@{\hskip 10pt}r@{\hskip 10pt}r r@{\hskip 10pt}r@{\hskip 10pt}r}
\toprule
& 
& \multicolumn{3}{c}{\textbf{Original Dataset}} 
& \multicolumn{3}{c}{\textbf{\textsc{StreamTrap}}} \\
\cmidrule(lr){3-5} \cmidrule(lr){6-8}
Dataset & Continent 
& \#\,Img & \#\,Cls & \#\,Traps
& \#\,Img & \#\,Cls & \#\,Traps \\
\midrule
\addlinespace[5pt]
Caltech Camera Traps~\cite{beery2018recognition}       & N.\ America & 243.1K & 22 & 140 &  37.6K & 13 & 15 \\ \addlinespace[10pt]
Idaho Camera Traps~\cite{IdahoCameraTraps}         & N.\ America &1535.7K & 62 & 276 &  43.6K & 10 & 15 \\ \addlinespace[10pt]
NA Camera Trap Img~\cite{tabak2019na}              & N.\ America &3280.8K & 59 & 551 & 964.9K & 20 & 52 \\ \addlinespace[10pt]
NZ Trail Cam.\ Animals~\cite{NZTrailcams}          & Oceania &2045.8K & 98 &2469 & 543.9K & 41 &139 \\ \addlinespace[10pt]
Orinoquía Camera Traps~\cite{velez2023evaluation}  & S.\ America & 112.2K & 58 &  49 &  20.1K & 16 &  6 \\ \addlinespace[10pt]
Snapshot APNR*~\cite{pardo2021snapshot}           & Africa & 228.4K & 60 & 148 & 102.2K & 19 & 40 \\ \addlinespace[10pt]
Snapshot Camdeboo*~\cite{pardo2021snapshot}        & Africa &  67.6K & 48 &  59 &  22.9K & 23 & 12 \\ \addlinespace[10pt]
Snapshot Enonkishu*~\cite{pardo2021snapshot}       & Africa & 132.8K & 48 &  83 &  59.7K & 23 & 13 \\ \addlinespace[10pt]
Snapshot Karoo*~\cite{pardo2021snapshot}           & Africa & 261.7K & 48 &  96 &  21.2K & 17 & 10 \\ \addlinespace[10pt]
Snapshot Kgalagadi*~\cite{pardo2021snapshot}       & Africa &  94.7K & 48 & 132 &  23.6K & 16 &  9 \\ \addlinespace[10pt]
Snapshot Kruger*~\cite{pardo2021snapshot}          & Africa &  59.6K & 59 & 124 &   5.8K & 11 &  3 \\ \addlinespace[10pt]
Snapshot Madikwe*~\cite{pardo2021snapshot}         & Africa & 277.5K & 69 & 136 & 122.1K & 28 & 28 \\ \addlinespace[10pt]
Snapshot Mt.\ Zebra*~\cite{pardo2021snapshot}      & Africa & 412.4K & 59 &  74 &  15.4K & 21 &  7 \\ \addlinespace[10pt]
Snapshot Pilanesberg*~\cite{pardo2021snapshot}     & Africa & 108.7K & 65 &  65 &  13.3K & 19 &  3 \\ \addlinespace[10pt]
Snapshot Ruaha*~\cite{pardo2021snapshot}           & Africa & 351.0K & 54 &  47 &   4.6K & 13 &  2 \\ \addlinespace[10pt]
Snapshot Serengeti~\cite{swanson2015serengeti}     & Africa &7178.4K & 61 & 225 &1308.4K & 45 &188 \\ \addlinespace[10pt]
Wellington Cam.\ Traps~\cite{anton2018monitoring}   & Oceania & 270.5K & 17 & 215 &   8.1K &  7 &  4 \\
\addlinespace[5pt]
\bottomrule
\end{tabular}%
}
\vspace{3mm}
\caption{Summary of dataset statistics from the selected original LILA BC sources and \textsc{StreamTrap} after applying the proposed data processing and filtering pipeline. Datasets marked with * indicate datasets grouped under the unified project \emph{Snapshot Safari 2024 Expansion}~\cite{pardo2021snapshot}. Image counts are reported in thousands (K) with one-decimal precision.}
\label{tab:dataset_summary}
\end{table*}
\clearpage

%% file: tables_eccv/camera_trap_index_new.tex
\begin{table}[H]
\centering
\scriptsize
\caption{Camera trap index reorganized by the new camera index. Each camera trap is assigned an index $C_i$ (e.g., $C_1$, $C_2$) used to reference camera traps in figures and tables throughout the paper.}
\vspace{-1mm}
\label{tab:cam_index_new}
\setlength{\tabcolsep}{8pt}
\renewcommand{\arraystretch}{1.4}
\noindent\resizebox{\textwidth}{!}{%
\begin{tabular}{llr@{\hskip 2.5em}llr}
\toprule
\textbf{Dataset} & \textbf{Camera Trap} & \textbf{Index C} &
\textbf{Dataset} & \textbf{Camera Trap} & \textbf{Index C} \\
\midrule
  Snapshot APNR & K082 & 1 & Snapshot Serengeti & L06 & 59 \\
  NA Camera Trap Img & lebec\_CA-19 & 2 & NZ Trail Cam. Animals & EFH\_HCAMH02 & 60 \\
  Snapshot Madikwe & G08 & 3 & NZ Trail Cam. Animals & EFD\_DCAMC05 & 61 \\
  NZ Trail Cam. Animals & EFH\_HCAMF12 & 4 & NZ Trail Cam. Animals & EFH\_HCAMC06 & 62 \\
  Snapshot Serengeti & G13 & 5 & Orinoquía Camera Traps & M04 & 63 \\
  NZ Trail Cam. Animals & PS1\_CAM7709 & 6 & NZ Trail Cam. Animals & EFH\_HCAMD10 & 64 \\
  NZ Trail Cam. Animals & EFD\_DCAMA02 & 7 & NZ Trail Cam. Animals & EFH\_HCAMA09 & 65 \\
  Snapshot Enonkishu & C02 & 8 & Snapshot Kgalagadi & KHOLA03 & 66 \\
  NZ Trail Cam. Animals & EFD\_DCAME01 & 9 & NZ Trail Cam. Animals & EFH\_HCAMC07 & 67 \\
  Snapshot Serengeti & M13 & 10 & Orinoquía Camera Traps & N29 & 68 \\
  Snapshot Serengeti & Q10 & 11 & NZ Trail Cam. Animals & EFH\_HCAMB10 & 69 \\
  Orinoquía Camera Traps & N04 & 12 & NZ Trail Cam. Animals & EFH\_HCAMD03 & 70 \\
  NZ Trail Cam. Animals & EFH\_HCAMB04 & 13 & Snapshot Mt. Zebra & F04 & 71 \\
  NZ Trail Cam. Animals & EFH\_HCAMC04 & 14 & Caltech Camera Trap & 70 & 72 \\
  Snapshot Kgalagadi & KHOLA08 & 15 & Snapshot Ruaha & M7 & 73 \\
  NZ Trail Cam. Animals & EFH\_HCAMH03 & 16 & Idaho Camera Traps & 85 & 74 \\
  Snapshot Camdeboo & B04 & 17 & Snapshot APNR & U43B & 75 \\
  NZ Trail Cam. Animals & EFH\_HCAMH04 & 18 & NZ Trail Cam. Animals & EFD\_DCAMG03 & 76 \\
  NZ Trail Cam. Animals & EFH\_HCAMC01 & 19 & Snapshot Enonkishu & B02 & 77 \\
  NZ Trail Cam. Animals & EFD\_DCAMB02 & 20 & Snapshot APNR & N1 & 78 \\
  Snapshot Mt. Zebra & E05 & 21 & Caltech Camera Trap & 100 & 79 \\
  Snapshot APNR & 13U & 22 & NA Camera Trap Img & lebec\_CA-36 & 80 \\
  Snapshot Camdeboo & E02 & 23 & Snapshot Pilanesberg & B04 & 81 \\
  NZ Trail Cam. Animals & EFD\_DCAMC03 & 24 & Snapshot APNR & K023 & 82 \\
  Snapshot Madikwe & H08 & 25 & Snapshot Madikwe & D07 & 83 \\
  Snapshot Serengeti & B04 & 26 & NA Camera Trap Img & lebec\_CA-21 & 84 \\
  Caltech Camera Trap & 88 & 27 & Snapshot Madikwe & A04 & 85 \\
  Snapshot Serengeti & L04 & 28 & NZ Trail Cam. Animals & EFD\_DCAMH07 & 86 \\
  NZ Trail Cam. Animals & EFD\_DCAMA08 & 29 & Snapshot APNR & K051 & 87 \\
  NA Camera Trap Img & lebec\_CA-16 & 30 & Snapshot Madikwe & B03 & 88 \\
  Snapshot Serengeti & D09 & 31 & Snapshot Karoo & A01 & 89 \\
  NZ Trail Cam. Animals & EFD\_DCAMF06 & 32 & Wellington Camera Traps & 031c & 90 \\
  NZ Trail Cam. Animals & EFD\_DCAME03 & 33 & Snapshot Serengeti & O13 & 91 \\
  NZ Trail Cam. Animals & EFH\_HCAMI01 & 34 & Caltech Camera Trap & 46 & 92 \\
  Snapshot Serengeti & Q09 & 35 & Idaho Camera Traps & 122 & 93 \\
  Orinoquía Camera Traps & M00 & 36 & NZ Trail Cam. Animals & EFH\_HCAME05 & 94 \\
  Snapshot Serengeti & E01 & 37 & NA Camera Trap Img & lebec\_CA-20 & 95 \\
  NZ Trail Cam. Animals & EFD\_DCAME09 & 38 & Snapshot Serengeti & Q11 & 96 \\
  Snapshot Serengeti & H11 & 39 & Snapshot Serengeti & I11 & 97 \\
  Snapshot Serengeti & K11 & 40 & Snapshot Serengeti & B03 & 98 \\
  NZ Trail Cam. Animals & EFD\_DCAME02 & 41 & NZ Trail Cam. Animals & EFH\_HCAMD07 & 99 \\
  NZ Trail Cam. Animals & EFH\_HCAME09 & 42 & Snapshot Madikwe & D03 & 100 \\
  Snapshot Serengeti & D12 & 43 & Snapshot Serengeti & O10 & 101 \\
  Snapshot Serengeti & U12 & 44 & Snapshot Madikwe & A05 & 102 \\
  NZ Trail Cam. Animals & EFH\_HCAMB09 & 45 & Snapshot Serengeti & I01 & 103 \\
  Snapshot Serengeti & E13 & 46 & Snapshot Serengeti & I12 & 104 \\
  NZ Trail Cam. Animals & EFH\_HCAME08 & 47 & NZ Trail Cam. Animals & EFH\_HCAMF02 & 105 \\
  NZ Trail Cam. Animals & EFD\_DCAMC07 & 48 & Snapshot Serengeti & L07 & 106 \\
  Snapshot Kgalagadi & KHOGC05 & 49 & Snapshot Serengeti & G07 & 107 \\
  NZ Trail Cam. Animals & EFH\_HCAMD08 & 50 & Snapshot Serengeti & O09 & 108 \\
  Orinoquía Camera Traps & N07 & 51 & Orinoquía Camera Traps & N25 & 109 \\
  NZ Trail Cam. Animals & EFH\_HCAMA05 & 52 & NZ Trail Cam. Animals & EFH\_HCAMC02 & 110 \\
  Snapshot Serengeti & L10 & 53 & NZ Trail Cam. Animals & EFD\_DCAMC04 & 111 \\
  NZ Trail Cam. Animals & EFH\_HCAMA06 & 54 & Snapshot Serengeti & C08 & 112 \\
  NZ Trail Cam. Animals & EFD\_DCAMB04 & 55 & Snapshot Serengeti & B10 & 113 \\
  Snapshot Serengeti & O04 & 56 & Snapshot Serengeti & E09 & 114 \\
  NZ Trail Cam. Animals & EFH\_HCAMB07 & 57 & NZ Trail Cam. Animals & EFH\_HCAMD05 & 115 \\
  NZ Trail Cam. Animals & EFH\_HCAMA02 & 58 & NZ Trail Cam. Animals & PS1\_CAM6213 & 116 \\
\bottomrule
\end{tabular}%
}
\end{table}

%% file: tables_eccv/vidhi_cam_list.tex
\setlength{\LTleft}{\fill}
\setlength{\LTright}{\fill}

\centering
\scriptsize
\renewcommand{\arraystretch}{1.1}

\begin{longtable}{@{}lrrr@{}}
\caption{\textsc{StreamTrap} overall camera trap statistics.}\label{tab:camera_summary}\\
\toprule
\textbf{Camera} & \textbf{\# Images} & \textbf{\# Intervals} & \textbf{\# Classes} \\
\midrule
\endfirsthead

\toprule
\textbf{Camera} & \textbf{\# Images} & \textbf{\# Intervals} & \textbf{\# Classes} \\
\midrule
\endhead

\midrule
\multicolumn{4}{r}{\emph{Continued on next page}} \\
\endfoot

\bottomrule
\endlastfoot

\rowcolor{gray!15} \multicolumn{4}{@{}l}{\textbf{Snapshot APNR}} \\
13U & 6,329 & 23 & 9 \\
6U & 2,081 & 10 & 8 \\
BOSP & 1,759 & 6 & 6 \\
JJ5 & 4,560 & 21 & 9 \\
JJ6 & 3,223 & 16 & 10 \\
JJ7 & 3,229 & 17 & 8 \\
JJP & 4,910 & 23 & 12 \\
K011 & 1,837 & 9 & 7 \\
K013 & 2,322 & 12 & 6 \\
K021 & 2,264 & 11 & 9 \\
K023 & 2,737 & 15 & 12 \\
K024 & 3,755 & 17 & 11 \\
K034 & 2,843 & 15 & 10 \\
K041 & 1,914 & 10 & 8 \\
K044 & 2,621 & 14 & 6 \\
K051 & 1,747 & 9 & 8 \\
K053 & 1,596 & 8 & 8 \\
K062 & 1,188 & 6 & 7 \\
K064 & 2,097 & 10 & 7 \\
K082 & 2,042 & 11 & 9 \\
KUMH & 1,297 & 6 & 6 \\
LL5 & 1,604 & 8 & 12 \\
N1 & 1,449 & 8 & 10 \\
N3 & 1,126 & 6 & 6 \\
N5 & 1,450 & 7 & 9 \\
RM4 & 1,640 & 8 & 7 \\
TB12 & 1,920 & 9 & 9 \\
TB14 & 3,548 & 16 & 10 \\
TB17 & 5,461 & 15 & 12 \\
TB26 & 5,768 & 17 & 10 \\
U23A & 1,429 & 7 & 8 \\
U24B & 1,560 & 8 & 7 \\
U43B & 1,743 & 9 & 6 \\
U63C & 3,108 & 14 & 8 \\
U64C & 2,536 & 12 & 12 \\
UMH8 & 1,846 & 9 & 9 \\
UMH9 & 3,691 & 18 & 10 \\
WB4 & 2,098 & 11 & 8 \\
WM & 2,201 & 11 & 6 \\
WS6 & 1,636 & 7 & 8 \\
\addlinespace[0.35em]
\rowcolor{gray!15} \multicolumn{4}{@{}l}{\textbf{Snapshot Camdeboo}} \\
A05 & 2,924 & 11 & 10 \\
A06 & 1,866 & 8 & 11 \\
B04 & 1,666 & 8 & 8 \\
B06 & 1,374 & 7 & 11 \\
B07 & 1,948 & 9 & 11 \\
C06 & 3,226 & 12 & 9 \\
C07 & 1,626 & 8 & 13 \\
D02 & 1,510 & 7 & 7 \\
D03 & 2,901 & 12 & 8 \\
D04 & 1,314 & 6 & 8 \\
D06 & 1,086 & 6 & 9 \\
E02 & 1,461 & 6 & 8 \\
\addlinespace[0.35em]
\rowcolor{gray!15} \multicolumn{4}{@{}l}{\textbf{Snapshot Enonkishu}} \\
B02 & 1,855 & 6 & 10 \\
B03 & 1,358 & 8 & 7 \\
B05 & 4,869 & 10 & 13 \\
B06 & 7,601 & 13 & 11 \\
C02 & 2,724 & 12 & 13 \\
C03 & 2,858 & 12 & 9 \\
C04 & 6,271 & 16 & 14 \\
C05 & 8,413 & 13 & 12 \\
C06 & 3,123 & 11 & 13 \\
D01 & 2,710 & 12 & 14 \\
D04 & 2,115 & 8 & 9 \\
D06 & 8,572 & 19 & 11 \\
E06 & 7,247 & 17 & 10 \\
\addlinespace[0.35em]
\rowcolor{gray!15} \multicolumn{4}{@{}l}{\textbf{Snapshot Karoo}} \\
A01 & 1,951 & 9 & 10 \\
A02 & 2,533 & 12 & 11 \\
B02 & 3,030 & 11 & 10 \\
B03 & 1,941 & 6 & 11 \\
C01 & 4,647 & 15 & 11 \\
D01 & 1,245 & 6 & 10 \\
D04 & 1,538 & 8 & 11 \\
E01 & 1,540 & 8 & 9 \\
E03 & 1,383 & 7 & 7 \\
F03 & 1,349 & 6 & 11 \\
\addlinespace[0.35em]
\rowcolor{gray!15} \multicolumn{4}{@{}l}{\textbf{Snapshot Kgalagadi}} \\
KHOGA04 & 1,568 & 7 & 7 \\
KHOGB03 & 3,612 & 10 & 9 \\
KHOGB04 & 1,847 & 7 & 7 \\
KHOGB06 & 3,118 & 11 & 9 \\
KHOGB07 & 3,406 & 11 & 9 \\
KHOGC05 & 1,561 & 7 & 9 \\
KHOLA03 & 2,158 & 7 & 7 \\
KHOLA05 & 2,805 & 10 & 5 \\
KHOLA08 & 3,533 & 10 & 5 \\
\addlinespace[0.35em]
\rowcolor{gray!15} \multicolumn{4}{@{}l}{\textbf{Snapshot Kruger}} \\
49 & 1,616 & 8 & 9 \\
7 & 2,679 & 10 & 11 \\
8 & 1,482 & 6 & 7 \\
\addlinespace[0.35em]
\rowcolor{gray!15} \multicolumn{4}{@{}l}{\textbf{Snapshot Madikwe}} \\
A03 & 2,064 & 9 & 7 \\
A04 & 1,886 & 9 & 9 \\
A05 & 3,536 & 12 & 10 \\
A06 & 1,603 & 8 & 6 \\
B03 & 1,928 & 8 & 8 \\
B05 & 5,837 & 13 & 9 \\
B06 & 1,444 & 7 & 9 \\
C04 & 5,867 & 15 & 10 \\
C06 & 6,992 & 14 & 8 \\
C07 & 2,522 & 11 & 8 \\
D03 & 1,743 & 9 & 8 \\
D04 & 1,304 & 7 & 8 \\
D07 & 3,699 & 15 & 9 \\
F03 & 1,254 & 6 & 5 \\
F04 & 1,515 & 7 & 7 \\
G04 & 2,107 & 10 & 10 \\
G05 & 1,481 & 7 & 7 \\
G08 & 1,757 & 8 & 8 \\
H08 & 1,556 & 9 & 10 \\
I04 & 1,771 & 9 & 12 \\
I05 & 2,250 & 9 & 7 \\
I06 & 2,987 & 10 & 8 \\
I08 & 1,347 & 7 & 10 \\
MAD01 & 10,599 & 11 & 15 \\
MAD03 & 21,236 & 10 & 14 \\
MAD04 & 10,580 & 11 & 15 \\
MAD05 & 4,488 & 11 & 12 \\
MAD06 & 16,721 & 11 & 14 \\
\addlinespace[0.35em]
\rowcolor{gray!15} \multicolumn{4}{@{}l}{\textbf{Snapshot Mountain Zebra}} \\
B04 & 1,230 & 6 & 7 \\
C08 & 1,691 & 9 & 11 \\
D03 & 2,472 & 11 & 17 \\
D06 & 1,402 & 7 & 6 \\
E05 & 1,891 & 9 & 11 \\
F04 & 1,864 & 9 & 8 \\
F05 & 4,870 & 13 & 6 \\
\addlinespace[0.35em]
\rowcolor{gray!15} \multicolumn{4}{@{}l}{\textbf{Snapshot Pilanesberg}} \\
B04 & 4,555 & 11 & 12 \\
C02 & 3,578 & 8 & 16 \\
D01 & 5,182 & 11 & 12 \\
\addlinespace[0.35em]
\rowcolor{gray!15} \multicolumn{4}{@{}l}{\textbf{Snapshot Ruaha}} \\
M2 & 2,824 & 8 & 12 \\
M7 & 1,814 & 6 & 8 \\
\addlinespace[0.35em]
\rowcolor{gray!15} \multicolumn{4}{@{}l}{\textbf{Snapshot Serengeti}} \\
B03 & 7,120 & 21 & 14 \\
B04 & 3,206 & 16 & 9 \\
B05 & 6,125 & 23 & 14 \\
B06 & 9,245 & 28 & 20 \\
B07 & 19,353 & 24 & 13 \\
B08 & 10,969 & 21 & 11 \\
B09 & 7,194 & 16 & 8 \\
B10 & 3,609 & 14 & 13 \\
B12 & 1,805 & 8 & 11 \\
B13 & 7,484 & 25 & 13 \\
C01 & 20,361 & 26 & 14 \\
C02 & 13,935 & 36 & 13 \\
C03 & 6,346 & 20 & 13 \\
C04 & 11,480 & 20 & 13 \\
C05 & 9,558 & 21 & 14 \\
C06 & 7,610 & 22 & 13 \\
C07 & 13,071 & 23 & 14 \\
C08 & 15,194 & 29 & 12 \\
C10 & 7,380 & 16 & 13 \\
C11 & 8,223 & 16 & 17 \\
C12 & 8,509 & 20 & 8 \\
C13 & 19,272 & 29 & 14 \\
D02 & 16,352 & 32 & 13 \\
D03 & 15,148 & 36 & 12 \\
D04 & 10,382 & 31 & 14 \\
D05 & 4,025 & 11 & 11 \\
D06 & 12,254 & 19 & 16 \\
D07 & 1,906 & 8 & 9 \\
D08 & 5,013 & 14 & 10 \\
D09 & 5,927 & 16 & 14 \\
D10 & 3,003 & 8 & 12 \\
D11 & 6,295 & 18 & 14 \\
D12 & 2,733 & 11 & 10 \\
D13 & 4,208 & 14 & 10 \\
E01 & 4,873 & 18 & 16 \\
E02 & 17,938 & 29 & 20 \\
E03 & 13,175 & 36 & 18 \\
E04 & 15,297 & 38 & 15 \\
E05 & 8,331 & 32 & 15 \\
E06 & 11,219 & 19 & 17 \\
E08 & 3,704 & 8 & 11 \\
E09 & 7,750 & 21 & 21 \\
E12 & 4,930 & 7 & 12 \\
E13 & 5,915 & 19 & 14 \\
F01 & 5,808 & 23 & 14 \\
F02 & 7,012 & 27 & 13 \\
F03 & 3,426 & 15 & 15 \\
F04 & 3,217 & 13 & 14 \\
F05 & 9,555 & 16 & 17 \\
F06 & 4,458 & 13 & 12 \\
F07 & 2,679 & 13 & 10 \\
F08 & 2,696 & 12 & 12 \\
F09 & 2,009 & 7 & 15 \\
F10 & 8,906 & 17 & 20 \\
F12 & 2,963 & 9 & 13 \\
F13 & 2,392 & 6 & 7 \\
G01 & 6,697 & 23 & 13 \\
G02 & 9,028 & 27 & 17 \\
G03 & 6,707 & 20 & 15 \\
G04 & 8,262 & 27 & 13 \\
G05 & 7,356 & 20 & 14 \\
G06 & 1,737 & 9 & 9 \\
G07 & 8,758 & 22 & 17 \\
G10 & 2,403 & 11 & 12 \\
G12 & 2,790 & 11 & 14 \\
G13 & 1,586 & 6 & 6 \\
H01 & 8,037 & 29 & 17 \\
H02 & 24,145 & 40 & 19 \\
H03 & 5,021 & 26 & 14 \\
H04 & 12,146 & 28 & 11 \\
H05 & 19,184 & 39 & 15 \\
H08 & 5,555 & 16 & 14 \\
H09 & 1,715 & 9 & 9 \\
H11 & 4,705 & 14 & 12 \\
H13 & 5,488 & 15 & 15 \\
I01 & 6,483 & 26 & 14 \\
I02 & 7,109 & 30 & 18 \\
I03 & 12,020 & 30 & 16 \\
I04 & 8,834 & 32 & 12 \\
I05 & 11,945 & 25 & 14 \\
I06 & 7,830 & 23 & 12 \\
I07 & 5,770 & 16 & 15 \\
I08 & 3,845 & 15 & 16 \\
I11 & 5,708 & 15 & 11 \\
I12 & 2,778 & 13 & 14 \\
I13 & 6,575 & 18 & 20 \\
J01 & 1,346 & 7 & 11 \\
J02 & 8,168 & 27 & 16 \\
J03 & 11,000 & 30 & 17 \\
J04 & 9,206 & 24 & 12 \\
J05 & 13,050 & 34 & 18 \\
J06 & 8,182 & 23 & 10 \\
J07 & 2,361 & 10 & 11 \\
J08 & 3,286 & 16 & 16 \\
J09 & 7,166 & 25 & 17 \\
J10 & 3,946 & 13 & 15 \\
J12 & 9,840 & 28 & 19 \\
J13 & 2,907 & 12 & 13 \\
K03 & 11,493 & 29 & 17 \\
K04 & 7,131 & 25 & 12 \\
K05 & 3,346 & 12 & 11 \\
K06 & 11,903 & 24 & 13 \\
K07 & 8,676 & 22 & 18 \\
K08 & 3,380 & 15 & 12 \\
K09 & 10,019 & 17 & 20 \\
K10 & 2,183 & 10 & 9 \\
K11 & 1,311 & 7 & 10 \\
K12 & 1,435 & 6 & 6 \\
K13 & 4,340 & 6 & 12 \\
L02 & 1,170 & 6 & 8 \\
L03 & 4,556 & 16 & 12 \\
L04 & 3,728 & 11 & 14 \\
L05 & 10,006 & 18 & 17 \\
L06 & 3,454 & 13 & 11 \\
L07 & 4,852 & 15 & 13 \\
L08 & 8,799 & 21 & 15 \\
L09 & 2,934 & 9 & 12 \\
L10 & 9,520 & 26 & 18 \\
L11 & 2,958 & 11 & 12 \\
M02 & 1,118 & 6 & 11 \\
M03 & 1,470 & 6 & 15 \\
M04 & 4,319 & 14 & 14 \\
M05 & 10,845 & 18 & 18 \\
M06 & 10,034 & 23 & 18 \\
M07 & 9,756 & 19 & 18 \\
M08 & 8,176 & 22 & 12 \\
M09 & 16,681 & 30 & 21 \\
M13 & 3,924 & 10 & 10 \\
N02 & 2,177 & 6 & 10 \\
N04 & 3,826 & 16 & 15 \\
N05 & 7,955 & 23 & 15 \\
N06 & 12,734 & 35 & 18 \\
N08 & 10,528 & 27 & 16 \\
N09 & 11,525 & 30 & 16 \\
N10 & 7,505 & 8 & 14 \\
N12 & 1,427 & 7 & 10 \\
O04 & 2,403 & 8 & 12 \\
O05 & 7,921 & 15 & 13 \\
O06 & 9,362 & 33 & 18 \\
O07 & 7,806 & 24 & 17 \\
O08 & 6,279 & 19 & 17 \\
O09 & 7,334 & 18 & 15 \\
O10 & 8,033 & 22 & 13 \\
O11 & 2,017 & 9 & 10 \\
O13 & 2,247 & 7 & 7 \\
P03 & 6,328 & 8 & 15 \\
P04 & 3,629 & 9 & 13 \\
P05 & 11,115 & 20 & 17 \\
P06 & 7,446 & 18 & 14 \\
P07 & 5,415 & 17 & 15 \\
P09 & 7,397 & 13 & 8 \\
P10 & 12,014 & 33 & 19 \\
P11 & 2,556 & 9 & 14 \\
P13 & 7,626 & 20 & 12 \\
Q04 & 2,188 & 8 & 12 \\
Q05 & 4,947 & 19 & 15 \\
Q06 & 5,622 & 17 & 12 \\
Q07 & 6,497 & 23 & 14 \\
Q08 & 7,799 & 25 & 14 \\
Q09 & 2,981 & 8 & 15 \\
Q10 & 1,807 & 9 & 10 \\
Q11 & 4,503 & 16 & 9 \\
Q12 & 4,731 & 14 & 11 \\
Q13 & 2,507 & 7 & 7 \\
R05 & 2,799 & 6 & 12 \\
R06 & 5,058 & 13 & 13 \\
R08 & 5,616 & 16 & 9 \\
R09 & 8,602 & 23 & 12 \\
R10 & 3,431 & 6 & 11 \\
R11 & 1,892 & 8 & 9 \\
R12 & 3,548 & 10 & 7 \\
S07 & 1,863 & 8 & 8 \\
S08 & 1,763 & 10 & 12 \\
S09 & 10,328 & 30 & 15 \\
S10 & 4,918 & 13 & 8 \\
S11 & 15,134 & 25 & 17 \\
S13 & 3,507 & 11 & 12 \\
T08 & 2,968 & 10 & 12 \\
T09 & 3,808 & 15 & 14 \\
T10 & 11,036 & 28 & 12 \\
T11 & 13,764 & 28 & 16 \\
T12 & 8,805 & 21 & 10 \\
T13 & 7,477 & 11 & 13 \\
U09 & 8,125 & 17 & 14 \\
U10 & 14,568 & 24 & 16 \\
U12 & 3,735 & 12 & 9 \\
U13 & 4,340 & 9 & 17 \\
V10 & 6,393 & 15 & 14 \\
\addlinespace[0.35em]
\rowcolor{gray!15} \multicolumn{4}{@{}l}{\textbf{North American Camera Trap Images}} \\
archbold\_FL-01 & 51,120 & 12 & 5 \\
archbold\_FL-04 & 21,313 & 16 & 6 \\
archbold\_FL-05 & 30,121 & 14 & 8 \\
archbold\_FL-09 & 48,065 & 10 & 8 \\
archbold\_FL-15 & 11,817 & 14 & 8 \\
archbold\_FL-16 & 80,565 & 14 & 8 \\
archbold\_FL-19 & 42,460 & 14 & 7 \\
archbold\_FL-20 & 22,038 & 14 & 6 \\
archbold\_FL-23 & 15,533 & 14 & 9 \\
archbold\_FL-26 & 13,765 & 13 & 7 \\
archbold\_FL-28 & 35,480 & 15 & 5 \\
archbold\_FL-30 & 26,237 & 14 & 5 \\
archbold\_FL-31 & 35,364 & 15 & 6 \\
archbold\_FL-37 & 16,756 & 11 & 8 \\
archbold\_FL-38 & 21,583 & 11 & 9 \\
archbold\_FL-39 & 5,973 & 10 & 8 \\
archbold\_FL-41 & 21,502 & 11 & 6 \\
archbold\_FL-42 & 8,783 & 9 & 5 \\
archbold\_FL-43 & 5,116 & 8 & 6 \\
archbold\_FL-44 & 2,776 & 9 & 6 \\
lebec\_CA-01 & 5,867 & 11 & 11 \\
lebec\_CA-02 & 8,827 & 16 & 11 \\
lebec\_CA-03 & 15,426 & 14 & 12 \\
lebec\_CA-04 & 14,661 & 17 & 12 \\
lebec\_CA-05 & 2,929 & 13 & 11 \\
lebec\_CA-06 & 12,450 & 18 & 12 \\
lebec\_CA-07 & 11,005 & 16 & 11 \\
lebec\_CA-09 & 5,597 & 15 & 11 \\
lebec\_CA-10 & 17,806 & 17 & 12 \\
lebec\_CA-11 & 13,471 & 14 & 12 \\
lebec\_CA-12 & 11,502 & 14 & 10 \\
lebec\_CA-13 & 8,598 & 14 & 9 \\
lebec\_CA-14 & 7,512 & 17 & 12 \\
lebec\_CA-16 & 3,811 & 12 & 12 \\
lebec\_CA-17 & 34,437 & 19 & 11 \\
lebec\_CA-18 & 30,290 & 18 & 10 \\
lebec\_CA-19 & 7,965 & 11 & 12 \\
lebec\_CA-20 & 20,483 & 14 & 11 \\
lebec\_CA-21 & 4,270 & 12 & 13 \\
lebec\_CA-24 & 9,246 & 15 & 12 \\
lebec\_CA-26 & 6,997 & 12 & 12 \\
lebec\_CA-27 & 13,616 & 15 & 12 \\
lebec\_CA-29 & 34,320 & 16 & 10 \\
lebec\_CA-30 & 41,674 & 18 & 12 \\
lebec\_CA-31 & 13,549 & 15 & 9 \\
lebec\_CA-32 & 3,730 & 9 & 9 \\
lebec\_CA-36 & 3,258 & 11 & 10 \\
lebec\_CA-37 & 27,415 & 15 & 10 \\
lebec\_CA-38 & 22,459 & 18 & 13 \\
lebec\_CA-40 & 4,054 & 11 & 14 \\
lebec\_CA-41 & 18,323 & 17 & 11 \\
lebec\_CA-45 & 12,950 & 10 & 10 \\
\addlinespace[0.35em]
\rowcolor{gray!15} \multicolumn{4}{@{}l}{\textbf{Idaho Camera Traps}} \\
100 & 4,865 & 8 & 7 \\
105 & 1,354 & 6 & 5 \\
109 & 1,356 & 6 & 6 \\
117 & 2,023 & 7 & 5 \\
118 & 3,191 & 10 & 6 \\
121 & 4,427 & 11 & 6 \\
122 & 4,319 & 10 & 6 \\
124 & 2,430 & 6 & 8 \\
84 & 2,545 & 7 & 7 \\
85 & 2,299 & 9 & 5 \\
86 & 3,120 & 10 & 7 \\
87 & 1,692 & 7 & 5 \\
89 & 4,562 & 10 & 5 \\
94 & 2,048 & 8 & 6 \\
98 & 3,380 & 7 & 5 \\
\addlinespace[0.35em]
\rowcolor{gray!15} \multicolumn{4}{@{}l}{\textbf{Caltech Camera Traps}} \\
100 & 2,721 & 10 & 7 \\
106 & 1,626 & 8 & 7 \\
114 & 4,896 & 10 & 9 \\
115 & 1,437 & 6 & 8 \\
120 & 1,208 & 6 & 5 \\
23 & 1,749 & 8 & 7 \\
33 & 1,614 & 7 & 7 \\
38 & 5,038 & 12 & 7 \\
43 & 1,903 & 8 & 7 \\
46 & 2,463 & 10 & 7 \\
57 & 3,951 & 11 & 7 \\
61 & 1,245 & 6 & 6 \\
70 & 2,568 & 10 & 7 \\
76 & 3,376 & 11 & 6 \\
88 & 1,755 & 8 & 8 \\
\addlinespace[0.35em]
\rowcolor{gray!15} \multicolumn{4}{@{}l}{\textbf{Orinoquía Camera Traps}} \\
M00 & 2,028 & 6 & 11 \\
M04 & 5,931 & 6 & 7 \\
N04 & 1,403 & 6 & 9 \\
N07 & 1,759 & 6 & 7 \\
N25 & 3,188 & 6 & 7 \\
N29 & 5,820 & 6 & 10 \\
\addlinespace[0.35em]
\rowcolor{gray!15} \multicolumn{4}{@{}l}{\textbf{Wellington Camera Traps}} \\
031c & 1,696 & 6 & 5 \\
046c & 1,834 & 6 & 5 \\
235 & 2,916 & 8 & 5 \\
324 & 1,697 & 6 & 5 \\
\addlinespace[0.35em]
\rowcolor{gray!15} \multicolumn{4}{@{}l}{\textbf{Trail Camera Images of New Zealand Animals}} \\
DWT\_BCAM419\_2022\_1 & 11,216 & 6 & 8 \\
DWT\_BCAM668\_2022\_1 & 2,873 & 6 & 7 \\
EFD\_DCAMA02 & 1,683 & 7 & 7 \\
EFD\_DCAMA04 & 2,969 & 13 & 10 \\
EFD\_DCAMA05 & 2,636 & 10 & 5 \\
EFD\_DCAMA06 & 2,736 & 11 & 8 \\
EFD\_DCAMA07 & 2,021 & 8 & 5 \\
EFD\_DCAMA08 & 2,268 & 9 & 9 \\
EFD\_DCAMA10 & 1,310 & 6 & 11 \\
EFD\_DCAMA11 & 1,605 & 7 & 9 \\
EFD\_DCAMB02 & 3,577 & 12 & 11 \\
EFD\_DCAMB03 & 2,276 & 8 & 6 \\
EFD\_DCAMB04 & 2,115 & 6 & 7 \\
EFD\_DCAMB06 & 2,645 & 8 & 6 \\
EFD\_DCAMB07 & 2,963 & 10 & 9 \\
EFD\_DCAMC02 & 2,422 & 9 & 8 \\
EFD\_DCAMC03 & 3,568 & 10 & 8 \\
EFD\_DCAMC04 & 3,378 & 7 & 8 \\
EFD\_DCAMC05 & 1,861 & 6 & 6 \\
EFD\_DCAMC07 & 1,700 & 7 & 8 \\
EFD\_DCAMC08 & 2,535 & 10 & 7 \\
EFD\_DCAMC09 & 1,609 & 7 & 6 \\
EFD\_DCAMC10 & 2,933 & 10 & 9 \\
EFD\_DCAMD02 & 2,648 & 10 & 10 \\
EFD\_DCAMD03 & 3,386 & 10 & 7 \\
EFD\_DCAMD04 & 3,238 & 11 & 7 \\
EFD\_DCAMD06 & 3,185 & 11 & 9 \\
EFD\_DCAMD07 & 3,164 & 11 & 11 \\
EFD\_DCAMD08 & 1,421 & 6 & 8 \\
EFD\_DCAMD10 & 2,408 & 9 & 6 \\
EFD\_DCAMD11 & 1,591 & 7 & 7 \\
EFD\_DCAME01 & 2,734 & 11 & 13 \\
EFD\_DCAME02 & 2,664 & 6 & 10 \\
EFD\_DCAME03 & 3,920 & 13 & 13 \\
EFD\_DCAME04 & 2,922 & 11 & 13 \\
EFD\_DCAME06 & 3,940 & 13 & 12 \\
EFD\_DCAME07 & 6,248 & 15 & 12 \\
EFD\_DCAME08 & 6,165 & 14 & 12 \\
EFD\_DCAME09 & 2,169 & 7 & 10 \\
EFD\_DCAMF02 & 1,414 & 6 & 8 \\
EFD\_DCAMF03 & 6,737 & 18 & 12 \\
EFD\_DCAMF06 & 3,950 & 13 & 10 \\
EFD\_DCAMF07 & 10,160 & 19 & 10 \\
EFD\_DCAMF08 & 4,313 & 15 & 14 \\
EFD\_DCAMF09 & 7,199 & 21 & 18 \\
EFD\_DCAMG03 & 2,576 & 11 & 6 \\
EFD\_DCAMH01 & 1,394 & 6 & 8 \\
EFD\_DCAMH02 & 1,456 & 6 & 6 \\
EFD\_DCAMH04 & 1,179 & 6 & 9 \\
EFD\_DCAMH06 & 3,924 & 12 & 9 \\
EFD\_DCAMH07 & 1,143 & 6 & 7 \\
EFD\_DCAMH08 & 3,220 & 9 & 5 \\
EFD\_DCAMH09 & 1,169 & 6 & 7 \\
EFH\_HCAMA02 & 4,543 & 8 & 10 \\
EFH\_HCAMA03 & 2,612 & 7 & 8 \\
EFH\_HCAMA05 & 6,954 & 10 & 7 \\
EFH\_HCAMA06 & 6,065 & 10 & 10 \\
EFH\_HCAMA07 & 5,350 & 13 & 7 \\
EFH\_HCAMA09 & 8,531 & 10 & 6 \\
EFH\_HCAMA10 & 5,290 & 8 & 10 \\
EFH\_HCAMA12 & 4,563 & 7 & 5 \\
EFH\_HCAMA13 & 1,908 & 7 & 6 \\
EFH\_HCAMB01 & 2,806 & 8 & 6 \\
EFH\_HCAMB02 & 2,898 & 9 & 5 \\
EFH\_HCAMB03 & 3,973 & 11 & 7 \\
EFH\_HCAMB04 & 5,767 & 13 & 5 \\
EFH\_HCAMB05 & 2,086 & 7 & 8 \\
EFH\_HCAMB07 & 6,675 & 14 & 10 \\
EFH\_HCAMB08 & 6,292 & 12 & 12 \\
EFH\_HCAMB09 & 4,124 & 12 & 11 \\
EFH\_HCAMB10 & 3,024 & 7 & 9 \\
EFH\_HCAMB11 & 1,667 & 6 & 7 \\
EFH\_HCAMC01 & 3,711 & 6 & 8 \\
EFH\_HCAMC02 & 5,822 & 11 & 7 \\
EFH\_HCAMC03 & 7,002 & 11 & 6 \\
EFH\_HCAMC04 & 7,950 & 12 & 10 \\
EFH\_HCAMC06 & 5,657 & 12 & 9 \\
EFH\_HCAMC07 & 6,136 & 13 & 10 \\
EFH\_HCAMC08 & 3,137 & 7 & 9 \\
EFH\_HCAMD01 & 3,048 & 6 & 8 \\
EFH\_HCAMD03 & 2,382 & 8 & 5 \\
EFH\_HCAMD04 & 6,641 & 12 & 12 \\
EFH\_HCAMD05 & 4,442 & 8 & 8 \\
EFH\_HCAMD06 & 3,400 & 10 & 11 \\
EFH\_HCAMD07 & 5,754 & 11 & 10 \\
EFH\_HCAMD08 & 4,996 & 13 & 12 \\
EFH\_HCAMD09 & 3,652 & 9 & 9 \\
EFH\_HCAMD10 & 2,178 & 6 & 9 \\
EFH\_HCAME02 & 11,502 & 16 & 10 \\
EFH\_HCAME03 & 11,811 & 15 & 9 \\
EFH\_HCAME04 & 8,765 & 15 & 12 \\
EFH\_HCAME05 & 4,372 & 10 & 6 \\
EFH\_HCAME06 & 3,061 & 11 & 8 \\
EFH\_HCAME07 & 5,113 & 14 & 8 \\
EFH\_HCAME08 & 5,117 & 12 & 8 \\
EFH\_HCAME09 & 3,758 & 11 & 8 \\
EFH\_HCAME10 & 4,730 & 14 & 11 \\
EFH\_HCAMF01 & 4,339 & 14 & 12 \\
EFH\_HCAMF02 & 9,138 & 19 & 17 \\
EFH\_HCAMF04 & 4,909 & 17 & 15 \\
EFH\_HCAMF07 & 2,288 & 7 & 11 \\
EFH\_HCAMF09 & 9,324 & 21 & 17 \\
EFH\_HCAMF11 & 1,918 & 9 & 13 \\
EFH\_HCAMF12 & 2,310 & 9 & 11 \\
EFH\_HCAMG07 & 4,249 & 13 & 15 \\
EFH\_HCAMG08 & 2,416 & 11 & 10 \\
EFH\_HCAMG10 & 2,298 & 9 & 9 \\
EFH\_HCAMG12 & 16,709 & 19 & 15 \\
EFH\_HCAMG13 & 4,044 & 16 & 16 \\
EFH\_HCAMG14 & 1,910 & 7 & 10 \\
EFH\_HCAMG15 & 3,676 & 12 & 14 \\
EFH\_HCAMH02 & 4,148 & 8 & 7 \\
EFH\_HCAMH03 & 3,239 & 7 & 11 \\
EFH\_HCAMH04 & 4,209 & 8 & 9 \\
EFH\_HCAMH05 & 5,167 & 12 & 7 \\
EFH\_HCAMH06 & 6,766 & 12 & 10 \\
EFH\_HCAMH07 & 2,429 & 7 & 5 \\
EFH\_HCAMH08 & 3,779 & 8 & 7 \\
EFH\_HCAMH09 & 2,356 & 6 & 9 \\
EFH\_HCAMI01 & 2,985 & 6 & 8 \\
EFH\_HCAMJ01 & 4,264 & 9 & 6 \\
EFH\_HCAMJ02 & 4,660 & 6 & 6 \\
EFH\_HCAMJ04 & 3,451 & 8 & 6 \\
EFH\_HCAMJ06 & 3,176 & 7 & 6 \\
HLO\_HPAT003 & 1,336 & 7 & 6 \\
HLO\_Moto2 & 2,257 & 9 & 6 \\
PS1\_CAM6213 & 1,808 & 6 & 5 \\
PS1\_CAM6604 & 1,698 & 7 & 5 \\
PS1\_CAM6803 & 3,567 & 7 & 6 \\
PS1\_CAM6808 & 1,701 & 7 & 5 \\
PS1\_CAM7312 & 1,545 & 6 & 6 \\
PS1\_CAM7709 & 2,103 & 6 & 5 \\
PS1\_CAM7811 & 2,955 & 6 & 6 \\
PS1\_CAM7816 & 4,961 & 7 & 6 \\
PS1\_CAM7920 & 3,099 & 8 & 5 \\
PS1\_CAM8001 & 1,823 & 6 & 7 \\
PS1\_CAM8008 & 1,808 & 6 & 8 \\
PS1\_CAM8306 & 1,342 & 6 & 5 \\
ZIO\_TRAILCAM01 & 5,918 & 6 & 5 \\
\addlinespace[0.35em]
\end{longtable}